\pdfoutput=1
\documentclass[twoside,twocolumn]{article}
\usepackage{blindtext} 

\usepackage[sc]{mathpazo} 
\usepackage[T1]{fontenc} 
\usepackage{microtype} 

\usepackage{cite}
\usepackage{amsmath,amssymb,amsfonts}
\usepackage{algorithmic}
\usepackage{graphicx}
\usepackage{extarrows}
\usepackage{multirow}
\usepackage{threeparttable}
\usepackage{array}
\usepackage{amssymb}
\usepackage{textcomp}
\usepackage{epstopdf}
\usepackage{threeparttable}
\usepackage{subfigure}

\usepackage[english]{babel} 

\usepackage[hmarginratio=1:1,left=18mm,right=18mm,top=32mm,columnsep=20pt]{geometry} 
\usepackage[hang, small,labelfont=bf,up,textfont=it,up]{caption} 
\usepackage{booktabs} 

\usepackage{lettrine} 

\usepackage{enumitem} 
\setlist[itemize]{noitemsep} 

\usepackage{abstract} 

\usepackage{titlesec} 
\renewcommand\thesection{\Roman{section}} 
\renewcommand\thesubsection{\roman{subsection}} 
\titleformat{\section}[block]{\large\scshape\centering}{\thesection.}{1em}{} 
\titleformat{\subsection}[block]{\large}{\thesubsection.}{1em}{} 

\usepackage{fancyhdr} 
\usepackage{titling} 

\usepackage{hyperref} 

\pretitle{\begin{center}\Huge\bfseries} 
\posttitle{\end{center}} 
\title{Asymmetric Deep Semantic Quantization for Image Retrieval} 
\author{%
\textsc{Zhan Yang$^1$, Osolo Ian Raymond$^1$, WuQing Sun$^1$, Jun Long$^{1,2}$}\thanks{Corresponding author. This work was supported in part by the Key Technology R\&D Program of Hunan Province (2018GK2052) and the Science and Technology Plan of Hunan (2016TP1003).} \\[1ex] 
\thanks{\copyright 2019 IEEE. Personal use of this material is permitted. Permission
from IEEE must be obtained for all other uses, in any current or future
media, including reprinting/republishing this material for advertising or
promotional purposes, creating new collective works, for resale or
redistribution to servers or lists, or reuse of any copyrighted
component of this work in other works. See \url{http://www.ieee.org/publications_standards/publications/rights/index.html} for more information}
\thanks{Accepted to IEEE ACCESS, DOI: 10.1109/ACCESS.2019.2920712}
\normalsize $^1$School of Computer Science and Engineering, Central South University, Changsha 410083, China \\ 
\normalsize $^2$Network Resources Management and Trust Evaluation Key Laboratory of Hunan Province \\ 
\normalsize \href{mailto:junlong@csu.edu.cn}{junlong@csu.edu.cn} 
}
\date{\today} 
\renewcommand{\maketitlehookd}{%
\begin{abstract}
Due to its fast retrieval and storage efficiency capabilities, hashing has been widely used in nearest neighbor retrieval tasks. By using deep learning based techniques, hashing can outperform non-learning based hashing technique in many applications. However, we argue that the current deep learning based hashing methods ignore some critical problems (e.g., the learned hash codes are not discriminative due to the hashing methods being unable to discover rich semantic information and the training strategy having difficulty optimizing the discrete binary codes). In this paper, we propose a novel image hashing method, termed as \textbf{\underline{A}}symmetric \textbf{\underline{D}}eep \textbf{\underline{S}}emantic \textbf{\underline{Q}}uantization (\textbf{ADSQ}). \textbf{ADSQ} is implemented using three stream frameworks, which consist of one \emph{LabelNet} and two \emph{ImgNets}. The \emph{LabelNet} leverages the power of three fully-connected layers, which are used to capture rich semantic information between image pairs. For the two \emph{ImgNets}, they each adopt the same convolutional neural network structure, but with different weights (i.e., asymmetric convolutional neural networks). The two \emph{ImgNets} are used to generate discriminative compact hash codes. Specifically, the function of the \emph{LabelNet} is to capture rich semantic information that is used to guide the two \emph{ImgNets} in minimizing the gap between the real-continuous features and the discrete binary codes. Furthermore, \textbf{ADSQ} can utilize the most critical semantic information to guide the feature learning process and consider the consistency of the common semantic space and Hamming space. Experimental results on three benchmarks (i.e., CIFAR-10, NUS-WIDE, and ImageNet) demonstrate that the proposed \textbf{ADSQ} can outperforms current state-of-the-art methods. 
\end{abstract}
}

\begin{document}
\maketitle


\section{Introduction}
\label{sec:introduction}
Over the last decade, the amount of image data available has increased exponentially. Finding ways to efficiently store and quickly search through this data has become a major challenge. Among all \textbf{N}earest \textbf{N}eighbor \textbf{S}earch (NNS)~\cite{bNNS} methods, hashing has been of considerable interest in many real-world applications in the image retrieval field due to its speedy search capabilities and low storage cost. In general, the basic hashing idea is to learn a mapping function $\{\mathcal{F}: \mathbb{R}^d\rightarrow \{0,1\}^k\}$, which map the original $d$-dimensional space (high-dimensional space) into $k$-bit Hamming space where the similarity is preserved. With the binary representation, the search speed for the data can be remarkably improved and the storage cost dramatically reduced. As a result of this, hashing techniques have become a popular tool for many image retrieval~\cite{b1,b2,b3,bdagh,b4} and text-image cross-model retrieval tasks~\cite{b5,b6}.

Hashing methods can be divided into two categories: data-independent hashing methods and data-dependent hashing methods. Data-independent hashing methods adopt random projections as hash functions to map the data points from the original high-dimension representation space into a low-dimension representation space (i.e, binary codes). In other words, data-independent hashing methods define a function that is not dependent on the data itself. Unfortunately, data-independent hashing methods need long hash codes to achieve satisfactory retrieval performance. In order to solve the limitation of data-independent hashing methods, recent works have shown that data-dependent hashing methods can achieve better performance with shorter hash codes. Data-dependent hashing methods, which can be further categorized into unsupervised and supervised methods, learn the hash function from training data points. Unsupervised methods are primarily measured by the use of distance metrics (e.g., Euclidean distance or cosine distance) of data point features~\cite{bisoh,bdgh,b8}. Therefore, in order to bridge the semantic gap, unlike the unsupervised hashing methods, supervised hashing methods utilize the semantic labels to boost the hash function quality. Many researchers currently focus on studying the supervised hashing methods~\cite{b4,b9,b11,b12,b15,bcos}. However, most of them map the original data point into binary codes by using the hand-crafted features, if the feature distribution of a dataset (large-scale dataset) is complex, the performance of these methods will decrease.

Fortunately, Deep Neural Networks (DNNs), especially Convolutional Neural Networks (CNNs) have been widely used in the computer vision field~\cite{b20} and have shown their powerful feature extraction capabilities. Inspired by this, some learning based hashing methods ~\cite{b10,b16,b17,b18} that adopt convolutional neural networks as the nonlinear hashing functions to enable end-to-end learning of learnable representations and hash codes, have demonstrated satisfactory retrieval performance on many benchmark datasets. Despite recent learning based hashing methods achieving significant progress in image retrieval, there are still some limitations to their usage, e.g., the label information is a simple construction of the similarity matrix, and does not make full use of the multiple label information of the data points~\cite{b32}. Taking the NUS-WIDE dataset as an example, there is an instance that is annotated with multiple labels, such as ``person'', ``tree'' and ``sea'', which can provide abundant semantic information and perfect similar relationship. As described in~\cite{bdseh}, a method named Deep Joint Semantic-Embedding Hashing (DSEH) that makes full use of multiple label information was proposed. This method can exploit the learned semantic correlation and hash codes in \emph{LabNet} as supervised information and transfer them to \emph{ImgNet}. However, there are still two limitations that should be addressed. Firstly, the real-continuous values will be converted by a relaxation scheme to the compact binary code. This is a mixed-integer optimization problem which results in an NP-hard optimization problem. To solve this issue, DSEH addresses the problem by quantizing the real-continuous values to compact binary values, which will cause a large quantization loss. Secondly, another limitation of the DSEH arises when measuring the similarity between each pair of image instances. This is measured by estimating the Hamming distance between the outputs of the same hash function. Because DSEH employs a symmetric structure (same structure with same weights, i.e, the same networks), this symmetric structure usually leads to the appearance of highly correlated bits in practice, which will degrade the retrieval performance. Intuitively, a pair of images with the same or different labels should not be seen as completely similar or dissimilar. Inspired by this, we use the asymmetric structure (same structure with different weights) to learn half of the codes which is capable of effectively decorrelating different bits, making the learned hash codes more informative.

To solve the above-mentioned challenges, this paper presents \textbf{\underline{A}}symmetric \textbf{\underline{D}}eep \textbf{\underline{S}}emantic \textbf{\underline{Q}}uantization (\textbf{ADSQ}) for efficient and effective image retrieval, which introduces a novel asymmetric training strategy for quantization and offering superior retrieval performance with three contributions detailed as below:

\begin{enumerate}
\item We develop a novel asymmetric framework for image retrieval, consisting of two ImgNets and one LabelNet. Two convolutional neural networks (i.e., ImgNets) are trained as different hash functions to generate compact binary codes for image pairs, and one fully-connected network (i.e., LabelNet) to capture abundant semantic correlation information from the image pair. The model effectively captures similarity relationships between the real-continuous features and binary hash codes, and can generate the discriminative compact hash codes.

\item Binary hash codes from training data points are learned with an iterative optimization strategy. Furthermore, based on the optimization scheme, an asymmetric loss between the binary-like codes and the learned discrete hash codes is imposed to reduce the quantization error.

\item Results from our experiments demonstrate that \textbf{ADSQ} outperforms several state-of-the-art methods for the task of image retrieval.
\end{enumerate}

The organization of the rest part is structured as follows. Section \ref{sec:Related_Work}  briefly introduces the related works on learning based hashing quantization. In Section \ref{sec:Method}, we formulate the problem and provide the details of the proposed training strategy. Section \ref{sec:EXPERIMENTS} shows the experimental results and Section \ref{sec:Conclusion_and_Future_Work} gives conclusion of this work.

\begin{figure*}
    \centering
    \includegraphics[scale=0.27]{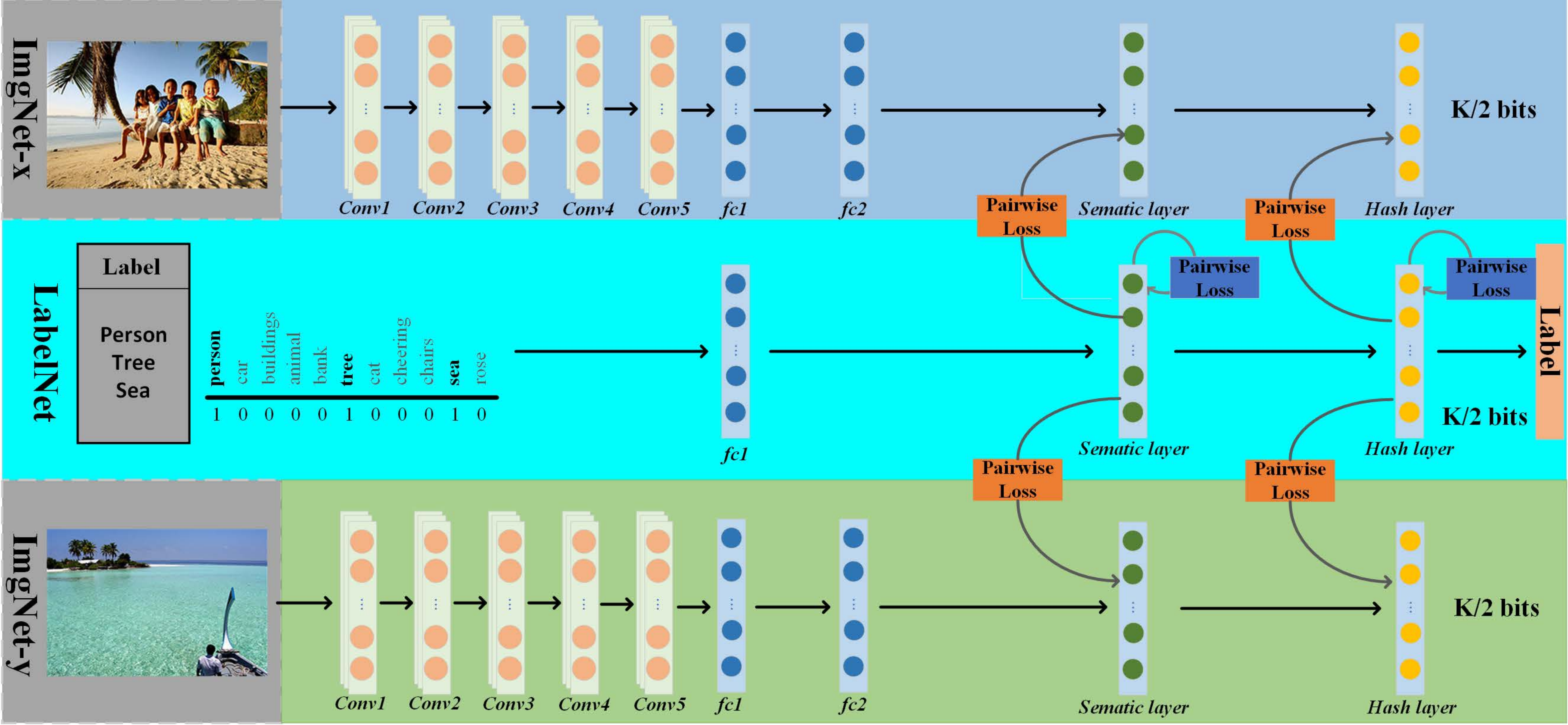}\\
    \caption{The proposed architecture for \textbf{\underline{A}}symmetric-Based \textbf{\underline{D}}eep \textbf{\underline{S}}emantic \textbf{\underline{Q}}uantization (\textbf{ADSQ}). \textbf{ADSQ} is an end-to-end deep learning framework which adopts two network streams: the first one (LabelNet) with three fully-connected layers (i.e., a common fully-connected layer, a semantic layer, and a hash layer.) is used for rich semantic feature extraction. The second consists of two asymmetric ImgNets with five convolutional layers and four fully-connected layers (i.e., two common fully-connected layers, a semantic layer, and a hash layer) that are used to generate the discriminative compact hash codes. During the ImgNets training process, we adopt an alternating strategy to learn the parameters of network and binary codes alternatingly. These codes are both guided by the supervised semantic information and the supervised binary-like representation generated by LabelNet. (Best viewed in color.)}\label{fig:F2}
  \end{figure*}

\section{Related Work}\label{sec:Related_Work}
By representing images as binary codes and taking advantage of fast query retrieval, the use of hashing techniques in image retrieval has attracted considerable attention. A comprehensive survey that covers the recent hashing techniques are provided in~\cite{b21}.

According to previous research, hashing methods can be roughly divided into two categories: data-independent and data-dependent methods. Spectral Hashing (SH)~\cite{b7} and Locality Sensitive Hashing (LSH) are two of the most common data-independent methods used. LSH aims to use several random projections such as hash functions to map the data points into a Hamming space~\cite{b22}. Some variants of LSH (e.g., kernel LSH~\cite{b23} and $p$-norm LSH~\cite{b24}) have been used to improve the performance of LSH. Unlike the data independent methods, the data-dependent methods attain more compact hash codes by combining datasets to achieve a better retrieval accuracy. Data dependent methods can be categorized into unsupervised and supervised methods. Unsupervised hashing methods aim to preserve the linkages among the unlabeled training data points. Typical examples include graph based hashing~\cite{bdgh,b29,b30}, minimize quantization error~\cite{bmqe}, and minimize reconstruction error~\cite{bisoh,b8,b25,b26,b28}. Supervised methods utilize the semantic labels or relevance information to improve the quality of hash codes. For example, Supervised Hashing with Kernels (KSH)~\cite{b9} and Supervised Discrete Hashing (SDH)~\cite{b15} generate binary hash codes by minimizing the Hamming distances across similar pairs of data points. Distortion Minimization Hashing (DMS)~\cite{b4}, Minimal Loss Hashing (MLH)~\cite{b11}, and Order Preserving Hashing (OPH)~\cite{b12} learn hash codes by minimizing the triplet loss based on similar pairs of data points. COlumn Sampling based Discrete Supervised Hashing (COSDISH)~\cite{bcos} learns the discrete hashing code from semantic information.

More recently, deep learning based hashing methods have shown superior performance by blending the powerful feature extraction of deep learning~\cite{b19,b20}. In particular, Convolutional Neural Network Hashing (CNNH)~\cite{b17} is a two-stage hashing method which learns hash codes and deep hash functions separately for image retrieval. Following this work, many learning based hashing techniques have been proposed, e.g., Weakly-shared Deep Transfer Networks (WDTN)~\cite{bwdtns} which can adequately mitigate the problem of insufficient image training data by bringing in rich labels from the text domain. Deep Semantic Ranking Hashing (DSRH)~\cite{b31} employs multilevel semantic ranking supervision to learn deep hash functions based on CNN which preserves the semantic structure if multi-label image. Deep Discrete Supervised Hashing (DDSH)~\cite{b1} utilizes pairwise supervised information to directly guide both discrete coding procedure and deep feature learning procedure and thus enhance the feedback between these two important procedures. Deep Supervised Hashing (DSH)~\cite{b32} utilizes a CNN architecture that takes pairs of images (similar of dissimilar) as training inputs and encourages the output of each image to approximate discrete values. Deep Ordinal Hashing (DOH)~\cite{bdoh} uses an effective spatial attention model to capture the local spatial information by selectively learning well-specified locations closely related to target objects. Generalized Deep Transfer Networks (DTNs)~\cite{bdtns} is a model which can learn the semantic knowledge from Web texts and then transfer it to images by the learned translator function when there is a lack of sufficient training data in the visual domain.  Network In Network Hashing (NINH)~\cite{b33} uses a ``one-stage'' supervised hashing method via a deep architecture that maps images to hash codes. Deep Supervised Discrete Hashing (DSDH)~\cite{b34} constrains the outputs of the last layer to be binary codes directly, and adopts an alternating minimization method to optimize the objective function by using the discrete cyclic coordinate descend method. Deep Joint Semantic-Embedding Hashing (DSEH)~\cite{bdseh} consists of \emph{LabNet} and \emph{ImgNet}. Specifically, \emph{LabNet} is explored to capture abundant semantic correlation between sample pairs and supervise \emph{ImgNet} from both semantic level and hash codes level.

However, even though DSEH captures abundant semantic correlation to indicate the accurate similarity relationship between samples, it is based on a shallow architecture which cannot effectively differentiate between the real-continuous features and discrete hash codes, because of their high degree of similarity. Therefore, in this paper, we propose a novel learning based hashing method, which can not only capture rich semantic correlation information, but also semantically associate the learned real-continuous features with the binary codes through an asymmetric network.

\section{Asymmetric Deep Semantic Quantization}\label{sec:Method}
In this section, we details the proposed \textbf{ADSQ} method. In order to attain robust image representations, the proposed \textbf{ADSQ} method that includes three stream frameworks, i.e., two ImgNets and a LabelNet. The two ImgNets which adopt the same convolutional neural network structure but with different weights, are used to generate discriminative compact hash codes. The LabelNet, which captures rich semantic correlation information, is used to guide the two ImgNets minimizing the quantization gap. As shown in~\cite{b19}, the top layers of the deep convolutional neural network can gradually extract global and more high-level representations. The details of our model are described in the following subsections.

\subsection{Notations and Problem Definition}

In this paper, we use boldface uppercase characters like $\boldsymbol{B}$ to denote a matrix, and vectors are denoted by boldface lowercase characters like $\boldsymbol{b}$. $\boldsymbol{B}_{ij}$ means the ($i,j$)-th element of $\boldsymbol{B}$. $\boldsymbol{B}^T$ is the transpose of $\boldsymbol{B}$, and the $\ell_2$-norm of a vector $\boldsymbol{b}\in \mathbb{R}^D$ is defined as $||\boldsymbol{b}||_2 = (\sum_{i=1}^{D}|b_i|^2)^{1/2}$. The Frobenius norm of a matrix $\boldsymbol{B}\in \mathbb{R}^{m\times n}$ as $||\boldsymbol{B}||_F^2=\sum_{i=1}^{m}\sum_{j=1}^{n}B_{ij}^2=\text{tr}[\boldsymbol{B}^T\boldsymbol{B}]$, while $\text{tr}[\boldsymbol{B}]$ is the trace of $\boldsymbol{B}$ if $\boldsymbol{B}$ is square. The symbol $\otimes$ denotes the element-wise product (i.e., Hadamard product). We use $\boldsymbol{1}$ to denote a vector with all elements being 1. The $sign(\cdot)$ is an element-wise $\boldsymbol{sign}$ function, and $sign(x)=1$ if $x\geq 0$, otherwise $sign(x)=-1$.

In similarity retrieval systems, we are given a training set $\mathcal{D}=\{\boldsymbol{d_i}\}_{i=1}^N$, $\boldsymbol{d_i}=\{\boldsymbol{x_i,y_i,l_i}\}$, where $\boldsymbol{x_i}\in \mathbb{R}^{1\times D}$ and $\boldsymbol{y_i}\in\mathbb{R}^{1\times D}$ to denote the feature vector of the $i$-th image in the first and second deep convolutional neural networks\footnote{Note that, although we use different symbols $\boldsymbol{x}$ and $\boldsymbol{y}$ to represent images, both of them denote the same training dataset.}, respectively. $\boldsymbol{l_i}=[l_{i1},l_{i2},...,l_{ic}]$ are the label annotations assigned to $\boldsymbol{d_i}$, where $c$ is the number of categories. Furthermore, for supervised learning based hashing methods, pairwise information can be used which is denoted by $\boldsymbol{S}=\{s_{ij}\}$\footnote{Note that one image may belong to multiple categories.}. If $s_{ij}=1$, it means that $\boldsymbol{x_i}$ and $\boldsymbol{y_j}$ are similar, while $s_{ij}=0$ implies that $\boldsymbol{x_i}$ and $\boldsymbol{y_j}$ are dissimilar. The goal of a learning based hashing method for quantization is to learn a quantizer $\mathcal{Q}:\boldsymbol{x}\rightarrow b_i\in\{-1,1\}^{K}$ from an input space $\mathbb{R}^D$ to Hamming space $\{-1,1\}^{K}$ with a deep neural network, where $K$ is the length of the binary codes. The similarity labels $\boldsymbol{S}=\{s_{ij}\}$ can be constructed from semantic labels of data points or relevance feedback in real retrieval systems.

For two binary hash codes $b_i$ and $b_j$, the similar relationship is defined according to a distance metric: $\boldsymbol{D}(b_i,b_j)$, where $\boldsymbol{D}(\cdot)$ is a distance metric function (e.g., Hamming distance or cosine distance). In this paper, the aim of our model is to learn two mapping functions $\mathcal{F}_{\boldsymbol{x}}$ and $\mathcal{F}_{\boldsymbol{y}}$ to map $\boldsymbol{X}$ and $\boldsymbol{Y}$ into the Hamming space $\boldsymbol{B}$: $b_i=sign(\mathcal{F}_{\boldsymbol{x}}(\boldsymbol{x_i}))\in\mathbb{R}^{K/2}$ and $b_j=sign(\mathcal{F}_{\boldsymbol{y}}(\boldsymbol{y_j}))\in\mathbb{R}^{K/2}$. For notation simplicity, we denote the length of the hash codes generated by each ImgNet from $K/2$, as $K$. Therefore, the length of the final hash codes is $2K$. We define the relationship between their Hamming distance $\boldsymbol{D}ist_H$ and inner product $\langle\cdot,\cdot\rangle$ can be calculated using: $\boldsymbol{D}ist_H=\frac{1}{2}(K-\langle b_i,b_j\rangle)$. Therefore, we can use the inner product operation to measure the similarity of two binary codes.

Given the pairwise similarity labels $\boldsymbol{S}=\{s_{ij}\}$, the logarithm Maximum a Posteriori (MAP) estimation of the hash codes $\boldsymbol{B}=[b_1,b_2,...,b_N]$ for all $N$ training points is:
\begin{equation}\label{eq:WL}
\begin{aligned}
  \text{log}P(\boldsymbol{B}|\boldsymbol{S})&\varpropto\text{log}P(\boldsymbol{S}|\boldsymbol{B})P(\boldsymbol{B})\\
  &=\sum_{S_{ij}}\text{log}P(s_{ij}|b_i,b_j)P(b_i,b_j),
\end{aligned}
\end{equation}
where $P(\boldsymbol{S}|\boldsymbol{B})$ denotes the likelihood function, and $P(\boldsymbol{B})$ is the prior distribution. For each pair, $P(s_{ij}|b_i,b_j)$ is the conditional probability of $s_{ij}$ given the pair of corresponding hash codes $[b_i,b_j]$, which is naturally defined by the binary distribution,
\begin{equation}\label{eq:logistic_function}
  P(s_{ij}|b_i,b_j)=\begin{cases}
                      \sigma(\langle b_i,b_j\rangle), &  s_{ij}=1\\
                      1-\sigma(\langle b_i,b_j\rangle), &  s_{ij}=0
                    \end{cases}
\end{equation}
where $\sigma(x)=1/(1+e^{-x})$, and $\langle b_i,b_j\rangle=\frac{1}{2}b_i^Tb_j$.

Similar to the hash layer, in the semantic layer, replace two real-continuous features $r_i$ and $r_j$ in~\eqref{eq:logistic_function}, the similar information between two real-continuous features can also be used in the same function. Therefore, the similarity probability of $r_i$ and $r_j$ can be expressed as binary distribution:
\begin{equation}\label{eq:continous_log_function}
  P(s_{ij}|r_i,r_j)=\begin{cases}
                      \sigma(\langle r_i,r_j\rangle), &  s_{ij}=1\\
                      1-\sigma(\langle r_i,r_j\rangle). &  s_{ij}=0
                    \end{cases}
\end{equation}

\subsection{LabelNet Training}
In this section, we have designed an end-to-end fully-connected neural network, named LabelNet, to bridge the semantic information at a more fine-grained level. Given a multiple label vector for instance, LabelNet extracts the semantic features layer-by-layer. Let $\mathcal{F}_l(\boldsymbol{l_i};W_l)$ denote embedding labels for label point $\boldsymbol{l_i}$, and $W_l$ denote the parameters of the LabelNet. Our goal is to maintain the similarity relationship between features and their corresponding hash codes. For LabelNet, the final loss can be defined as follows:
\begin{equation}\label{eq:labelloss}
  \begin{aligned}
  \underset{W_l}{\min}\ \mathcal{L}^l=&\ \ \alpha\mathcal{J}_1+\beta\mathcal{J}_2+\gamma\mathcal{J}_3+\delta\mathcal{J}_4\\
  =&-\alpha\sum_{s_{ij}\in\boldsymbol{S}}(s_{ij}\boldsymbol{\Lambda}_{ij}^l-\log(1+e^{\boldsymbol{\Lambda}_{ij}^l}))\\
  &-\beta\sum_{s_{ij}\in\boldsymbol{S}}(s_{ij}\boldsymbol{\Theta}_{ij}^l-\log(1+e^{\boldsymbol{\Theta}_{ij}^l}))\\
  &+\gamma\sum_{s_{ij}\in\boldsymbol{S}}(||\boldsymbol{\omega}^l_{i}-\boldsymbol{1}||_1+||\boldsymbol{\omega}^l_{j}-\boldsymbol{1}||_1)\\
  &+\delta\sum_{i=1}^N||\tilde{\boldsymbol{L}}-\boldsymbol{L}||_F^2,
  \end{aligned}
\end{equation}
where $\boldsymbol{\Lambda}_{ij}^l=\frac{1}{2}(r_i^l)^T(r_j^l)$, $\boldsymbol{\Theta}_{ij}^l=\frac{1}{2}(\omega_i^l)^T(\omega_j^l)$. $r_i^l$ denotes the semantic representation. $\omega^l$ represents the binary-like codes which are obtained by the output of the LabelNet and $\boldsymbol{\tilde{L}}=[\tilde{l}_1,\tilde{l}_2,...,\tilde{l}_i]$, $\tilde{l}_i=(W^l)^T\omega_i^l+b_i^l$ are the predicted labels of output, $\boldsymbol{L}$ is the true label. $\alpha,\ \beta,\ \gamma,\ \delta$ are hyper-parameters. In~\eqref{eq:labelloss}, $\mathcal{J}_1$ and $\mathcal{J}_2$ are the intra-pairwise loss terms. $\mathcal{J}_1$ is used to preserve the similarity information between semantic features in the semantic space whereas $\mathcal{J}_2$ is used to preserve the similarity between hashing features in the Hamming space. $\mathcal{J}_3$ is the binary regularization (i.e, to promote the hash code discretization), and $\mathcal{J}_4$ is to maintain the classification loss between the true label and the predicted label.

\subsection{ImgNet Training}
The image framework of the proposed method is shown in Fig~\ref{fig:F2}. As can be seen, we designed two end-to-end networks, named ImgNet-$\boldsymbol{x}$ and ImgNet-$\boldsymbol{y}$, which can map the features of an image into binary codes. These ImgNets are guided by LabelNet using the semantic features and the learned hash codes. $\mathcal{F}_x(\boldsymbol{x}_i,W_x)$ represents the output of the $i$-th image in the last layer of the ImgNet-$\boldsymbol{x}$, where $W_x$ stands for the parameters of the network. Similarly, we can obtain the output $\mathcal{F}_y(\boldsymbol{y}_j,W_y)$ corresponding to the $j$-th image using the parameters $W_y$ in the ImgNet-$\boldsymbol{y}$. In order to learn the optimal hash codes which can preserve the similarity information between the learned binary codes and the real-value features, one common way is to minimize the Frobenius norm between the similarity information and the inner product of the learned binary codes and the real-value features:
\begin{equation}\label{eq:imgL2}
  \min||\boldsymbol{I}^T\boldsymbol{B}^\kappa-K\boldsymbol{S}||_F^2,
\end{equation}
where $\boldsymbol{I}$ which denotes $sign(\mathcal{F}_\kappa(\kappa,W_\kappa)),\kappa=\boldsymbol{x},\boldsymbol{y}$. $\boldsymbol{B}^\kappa$ represents the learned binary codes. $K$ is the length of hash codes. $\boldsymbol{S}$ is the pairwise supervised information.

However, there exists a problem in the formulation in~\eqref{eq:imgL2}, it is difficult to implement a back-propagation (BP) algorithm for the gradient with respect to $\boldsymbol{I}$ due to their gradients always being zero. Hence, in this paper, we adopt $\tanh(\cdot)$ to approximate the threshold function $sign(\cdot)$. Thus, Equation~\eqref{eq:imgL2} is transformed into:
\begin{equation}\label{eq:imgL2Tanh}
  \min||{\boldsymbol{\tilde{I}}}^T\boldsymbol{B}^\kappa-K\boldsymbol{S}||_F^2,
\end{equation}
where $\boldsymbol{\tilde{I}}$ denotes $\tanh(\mathcal{F}_\kappa(\kappa,W_\kappa)),\kappa=\boldsymbol{x},\boldsymbol{y}$. For ImgNet, the final loss can be defined as follows:
\begin{equation}\label{eq:ImgLoss}
  \begin{aligned}
  \underset{\boldsymbol{B^\kappa},W_\kappa}{\min}\mathcal{L}^\kappa=&\ \ \alpha\mathcal{J}_1+\beta\mathcal{J}_2+\eta\mathcal{J}_3+\nu\mathcal{J}_4+\mathcal{A}\\
  =&-\alpha\sum_{s_{ij}\in\boldsymbol{S}}(s_{ij}\boldsymbol{\Lambda}_{ij}^\kappa-\log(1+e^{\boldsymbol{\Lambda}_{ij}^\kappa}))\\
  &-\beta\sum_{s_{ij}\in\boldsymbol{S}}(s_{ij}\boldsymbol{\Theta}_{ij}^\kappa-\log(1+e^{\boldsymbol{\Theta}_{ij}^\kappa}))\\
  &+\eta\ ||\boldsymbol{\tilde{I}}-\boldsymbol{B}^\kappa||_F^2\\
  &+\nu\ ||\boldsymbol{\tilde{I}}^T\boldsymbol{1}||_F^2\\
  &+||{\boldsymbol{\tilde{I}}}^T\boldsymbol{B}^\kappa-K\boldsymbol{S}||_F^2\\
  & s.t.\ \ \ \ \kappa=\boldsymbol{x},\boldsymbol{y}, \ \ \boldsymbol{B}^\kappa\in\{-1,+1\}^{n\times K},
  \end{aligned}
\end{equation}
where $\boldsymbol{\Lambda}_{ij}^\kappa=\frac{1}{2}(r_i^l)^T(r_j^\kappa)$, and $\boldsymbol{\Theta}_{ij}^\kappa=\frac{1}{2}(\omega_i^l)^T(\omega_j^\kappa)$, $r_i^l$ and $r_k^\kappa$ are semantic representations from LabelNet and ImgNets, respectively. $\omega^\kappa$ represents the binary-like codes which are obtained from the output of the ImgNets. $\alpha,\ \beta,\ \eta,\ \nu$ are the hyper-parameters. In~\eqref{eq:ImgLoss}, $\mathcal{J}_1$ and $\mathcal{J}_2$ are two negative-log likelihood functions (a.k.a. $\mathcal{J}_1$ and $\mathcal{J}_2$ exploit the inter-class and intra-class information). Note that although $\mathcal{J}_1$ and $\mathcal{J}_2$ in~\eqref{eq:labelloss} and~\eqref{eq:ImgLoss} are similar they represent different meanings. As such, we use the supervised features $r^l_i$ and $\omega_i^l$ which are learned from the LabelNet to guide the training of the asymmetric ImgNets. The relevance can be established using the LabelNet. Therefore, the semantic information can be fully utilized. $\mathcal{J}_3$ is the approximation loss between binary-like codes and hash codes. Note that, $\mathcal{J}_4$ makes a balance for each bit, which encourages the number of negative and positive numbers ($\pm1$) to be approximately similar among all data points (i.e., $\mathcal{J}_4$ is used to maximize the information provided by each bit)~\cite{bj4}. $\mathcal{A}$ is the asymmetric term, this term is used to exploit the semantic information between the binary code and real-value data.

\subsection{Optimization}
In this section, we introduce the training strategy. Firstly, we randomly initialize LabelNet and train it until~\eqref{eq:labelloss} converges. Secondly, we use the semantic representations and binary-like codes generated by LabelNet to guide the ImgNet training. Finally, the training procedure is repeated for LabelNet and ImgNet until convergence. Here, we only present the training detail for problem~\eqref{eq:ImgLoss} since problem~\eqref{eq:labelloss} can be easily adapted by using stochastic gradient descent with a back-propagation algorithm. Hence, we optimize the problem~\eqref{eq:ImgLoss} through iterative optimization. Specifically, in each iteration we learn one variable with the other fixed, and so on.

\textbf{$\boldsymbol{W_\kappa}$-step}: Fixing $B^\kappa$ to solve $W_\kappa$, then the objective problem can be transformed into:
\begin{equation}\label{eq:w-step-1}
  \begin{aligned}
  \underset{W_\kappa}{\min}\ \ &||\tanh(\mathcal{F}_\kappa(\kappa_i,W_\kappa))^T\boldsymbol{B}^\kappa-K\boldsymbol{S}||_F^2\\
  &-\alpha\sum_{s_{ij}\in\boldsymbol{S}}(s_{ij}\boldsymbol{\Lambda}_{ij}^\kappa-\log(1+e^{\boldsymbol{\Lambda}_{ij}^\kappa}))\\
  &-\beta\sum_{s_{ij}\in\boldsymbol{S}}(s_{ij}\boldsymbol{\Theta}_{ij}^\kappa-\log(1+e^{\boldsymbol{\Theta}_{ij}^\kappa}))\\
  &+\eta\ ||\tanh(\mathcal{F}_\kappa(\kappa_i,W_\kappa))-\boldsymbol{B}^\kappa||_F^2\\
  &+\nu\ ||\tanh(\mathcal{F}_\kappa(\kappa_i,W_\kappa))^T\boldsymbol{1}||_F^2.
  \end{aligned}
\end{equation}
Then we use the Back-Propagation (BP) algorithm to update $W_\kappa$. For the sake of simplicity, we define $\boldsymbol{v}_i=\mathcal{F}_\kappa(\kappa_i,W_\kappa)$ and $\boldsymbol{u}_i=\tanh(\mathcal{F}_\kappa(\kappa_i,W_\kappa))$. Then we can compute the gradient of $\boldsymbol{v}_i$ as follows:
\begin{equation}\label{eq:w-step-2}
  \begin{aligned}
    \frac{\partial \mathcal{L}^\kappa}{\partial\boldsymbol{v}_i}&= \bigg\{2\sum_{s_{ij}\in\boldsymbol{S}}[(b_j^T\boldsymbol{u}_i-KS_{ij})b_j+\frac{\alpha}{2}(\sigma(\Lambda_{ij})r_j^l-S_{ij}r_j^l)\\
    &+\frac{\beta}{2}(\sigma(\Theta_{ij})\omega_j^l-S_{ij}\omega_j^l)]+2\eta(\boldsymbol{u}_i-b_i)+2\nu\boldsymbol{U}^T\boldsymbol{1}\bigg\}\\
    &\otimes(1-\boldsymbol{u}_i^2),
  \end{aligned}
\end{equation}
where $r_j^l$ and $\omega_j^l$ are semantic representations and Hamming representations generated from LabelNet, respectively. $\boldsymbol{U}=[\boldsymbol{u}_{1},\boldsymbol{u}_{2},...,\boldsymbol{u}_{i}]$, symbol $\otimes$ denotes the Hadamard product. After getting the gradient $\frac{\partial\mathcal{L}^\kappa}{\partial\boldsymbol{v}_i}$, the chain rule is used to obtain $\frac{\partial\mathcal{L}^\kappa}{\partial W_\kappa}$, and $W_\kappa$ is updated by using the standard BP algorithm.

\textbf{$\boldsymbol{B^\kappa}$-step}: Fixing $W_\kappa$ to solve $B^\kappa$, then the objective problem can be transformed into:
\begin{equation}\label{eq:b-step-1}
  \begin{aligned}
  \underset{B^\kappa}{\min}\ \ &||\boldsymbol{U}{\boldsymbol{B}^\kappa}^T-K\boldsymbol{S}||_F^2+\eta(||\boldsymbol{U}-\boldsymbol{B}^\kappa||_F^2)\\
  &s.t.\ \ \boldsymbol{B}_i^\kappa\in\{-1,+1\}^{n\times K},
  \end{aligned}
\end{equation}
where $\boldsymbol{U}=[\boldsymbol{u}_{1},\boldsymbol{u}_{2},...,\boldsymbol{u}_{i}]$, Then~\eqref{eq:b-step-1} can be rewrote as:
\begin{equation}\label{eq:b-step-2}
  \begin{aligned}
  \underset{B^\kappa}{\min}\ \ &\text{tr}[\boldsymbol{B}^\kappa\boldsymbol{P}]+||\boldsymbol{B}^\kappa\boldsymbol{U}^T||_F^2+c\\
  &s.t.\ \ \boldsymbol{B}_i^\kappa\in\{-1,+1\}^{n\times K},
  \end{aligned}
\end{equation}
where $c$ means a constant value and $\boldsymbol{P}=-2K\boldsymbol{S}^T\boldsymbol{U}-2\eta\boldsymbol{U}$. According to~\cite{b3}, we can update $\boldsymbol{B}^\kappa$ bit by bit. In other words, we update one column of $\boldsymbol{B}^\kappa$ with other columns fixed. Let $\boldsymbol{B}^\kappa_{*c}$ denote the $c$-th column and $\boldsymbol{\tilde{B}}^\kappa_c$ denote the remaining columns in $\boldsymbol{B}^\kappa$. Let $\boldsymbol{U}_{*c}$ denote the $c$-th column of $\boldsymbol{U}$ and $\boldsymbol{\tilde{U}}_c$ denote the matrix of $\boldsymbol{U}$ excluding $\boldsymbol{U}_{*c}$. Let $\boldsymbol{P}_{*c}$ denote the $c$-th column of $\boldsymbol{P}$ and $\boldsymbol{\tilde{P}}_c$ denote the remaining columns in $\boldsymbol{P}$. Then~\eqref{eq:b-step-2} can be rewrote as:
\begin{equation}\label{eq:b-step-3}
  \begin{aligned}
  \underset{B^\kappa_{*c}}{\min}\ \ &\text{tr}(\boldsymbol{B}^\kappa_{*c}[2\boldsymbol{U}^T_{*c}\boldsymbol{\tilde{U}}_c{\boldsymbol{\tilde{B}}^\kappa_c}^T+\boldsymbol{P}_{*c}^T])+c\\
  &s.t.\ \ \boldsymbol{B}^\kappa\in\{-1,+1\}^{n\times K}.
  \end{aligned}
\end{equation}

The optimal solution of~\eqref{eq:b-step-3} can be found as follows:
\begin{equation}\label{eq:b-f}
  \boldsymbol{B}^\kappa_{*c}=-sign(2\boldsymbol{\tilde{B}}^\kappa_c\boldsymbol{\tilde{U}}_c^T\boldsymbol{U}_{*c}+\boldsymbol{P}_{*c}),
\end{equation}

Equation~\eqref{eq:b-f} can be used repeatedly until all columns are updated.

\begin{table}
\label{Al:Algorithm 1}
\setlength{\tabcolsep}{3pt}
\begin{tabular}{p{240pt}}
\hline
\specialrule{0em}{2pt}{2pt}
\textbf{Algorithm 1} \textbf{\underline{A}}symmetric \textbf{\underline{D}}eep \textbf{\underline{S}}emantic \textbf{\underline{Q}}uantization (\textbf{ADSQ}).\\
\specialrule{0em}{2pt}{2pt}
\hline
\specialrule{0em}{2pt}{2pt}
\textbf{Input} Training set $(\boldsymbol{X},\boldsymbol{Y},\boldsymbol{L})$; similarity matrix $\boldsymbol{S}\in\{-1,+1\}^{n\times n}$; hash code length $K$.\\
\specialrule{0em}{2pt}{2pt}
\textbf{Output} Parameters $\boldsymbol{W}_x$ of Hashing functions $\mathcal{F}_x$ and parameters $\boldsymbol{W}_y$ of $\mathcal{F}_y$.\\
\specialrule{0em}{2pt}{2pt}
\textbf{Initialization} Network parameters: $W_\kappa$, $W_l$, where $\kappa=\boldsymbol{x},\ \boldsymbol{y}$. Hyper-parameters: $\alpha$, $\beta$, $\gamma$, $\nu$, and $\eta$. Iteration number: $T^l$, $T^v$. Learning rate: $\mu$.\\
\specialrule{0em}{2pt}{2pt}
\textbf{repeat}\\
\specialrule{0em}{1pt}{1pt}
1.\ \textbf{for} $t=1:T^l$ epoch \textbf{do}\\
\specialrule{0em}{1pt}{1pt}
2.\ \ \ Update $W_l$ by standard BP algorithm:\\
\specialrule{0em}{1pt}{1pt}
3.\ \ \ $W_l\leftarrow W_l-\mu\nabla_{W_l}\mathcal{L}^l$ according to~\eqref{eq:labelloss}.\\
\specialrule{0em}{1pt}{1pt}
4.\ \textbf{end for}\\
\specialrule{0em}{1pt}{1pt}
1.\ \textbf{for} $t=1:T^v$ epoch \textbf{do}\\
\specialrule{0em}{1pt}{1pt}
2.\ \ \ Update $\boldsymbol{W}_\kappa$: Fixing $\boldsymbol{B}^\kappa$ to solve $\boldsymbol{W}_\kappa$ using standard BP algorithm\\
\ \ \ \ \ \ according to~\eqref{eq:w-step-2}, $\kappa=\boldsymbol{x},\ \boldsymbol{y}$.\\
\specialrule{0em}{1pt}{1pt}
3.\ \ \ Update $\boldsymbol{B}^\kappa$: Fixing $\boldsymbol{W}_\kappa$ to solve $\boldsymbol{B}^\kappa$ according to~\eqref{eq:b-f}, $\kappa=\boldsymbol{x},\ \boldsymbol{y}$.\\
\specialrule{0em}{1pt}{1pt}
5.\ \textbf{end for}\\
\specialrule{0em}{1pt}{1pt}
\textbf{until} convergence\\
\specialrule{0em}{1pt}{1pt}
\hline
\end{tabular}
\end{table}

\subsection{Out-of-Sample Extension}
When $\boldsymbol{W}_\kappa$ are learned, the asymmetric hash functions corresponding to the two asymmetric ImgNets are obtained. For example, given a new instance $x_q\notin\mathcal{X}$, we directly use it as the input of the \textbf{ADSQ} model, each model only needs to output $K/2$-bit hash codes, which are $b_{i}^q=sign(\mathcal{F}_{x}(\boldsymbol{x_q},W_x)\in\mathbb{R}^{K/2}$ and $b_{j}^q=sign(\mathcal{F}_y(\boldsymbol{x_q},W_{y}))\in\mathbb{R}^{K/2}$, respectively. Therefore, we concatenate the two $K/2$-bit binary codes to obtain the final hash codes:
\begin{equation}\label{eq:out_of_sample_hashing}
  b_q=concat[b_{i}^q,b_{j}^q]\in\mathbb{R}^K.
\end{equation}

\section{Experiments}\label{sec:EXPERIMENTS}
In this section, we evaluate the proposed \textbf{ADSQ} hashing with comparisons to the state-of-the-art methods~\cite{b7,b8,b15,b9,b16,b17,b33,b51,b34,bdseh} on three benchmark datasets.

\subsection{Datasets and Settings}
\noindent\textbf{CIFAR-10} is a standard dataset contains 60,000 images with 10 categories including ``truck'', ``airplane'', ``ship'', ``automobile'', ``horse'', ``bird'', ``cat'', ``deer'', ``frog'', ``dog''. We randomly selected 100 images per class as query set (totally 1,000 images), 500 images per class as the training set (totally 5,000 images). The rest of the images are used as the database.

\noindent\textbf{NUS-WIDE}~\cite{b57} is a multi-label image dataset contains 269,648 images collected from Flickr.com with 81 ground truth concepts. Following~\cite{b17} and~\cite{b34}, we filter 21 most common classes.  The 100 images per class are selected as the query set (totally 2,100 images), and 500 images per class are selected as the training set (totally 10,500 images). The rest of the images are used as the database. Two images are treated as similar if they share at least 1 common label. Otherwise, they are considered to be dissimilar.

\noindent\textbf{ImageNet}~\cite{b58} is a benchmark dataset contains over 1.2M images. It is a single-label dataset, where each image is labeled by one of 1,000 classes. Following~\cite{b3} and~\cite{b53}, we randomly select 100 classes, and randomly select 50 images per class as the query set (totally 5,000 images), 100 images per class as the training set (totally 10,000 images).

\subsection{Baselines}
We compared our proposed \textbf{ADSQ} method with ten state-of-the-art hashing methods, including: unsupervised hashing methods, supervised hashing methods, learning based hashing methods and semantic supervised learning based method. The unsupervised hashing methods used include: \textbf{SH}~\cite{b7}, \textbf{ITQ}~\cite{b8}, and supervised hashing methods: \textbf{SDH}~\cite{b15}, \textbf{KSH}~\cite{b9}. The learning based hashing methods used include \textbf{DPSH}~\cite{b51}, \textbf{DHN}~\cite{b16}, \textbf{CNNH}~\cite{b17}, \textbf{DNNH}~\cite{b33}, \textbf{DSDH}~\cite{b34}. The semantic supervised learning based method chosen was \textbf{DSEH}~\cite{bdseh}. We adopted $\text{DeCAF}_7$ features~\cite{b60} for non-deep learning based methods. For the deep learning based methods, the AlexNet~\cite{b19} or CNN-F~\cite{b59} network is used for comparison.

In this paper, we adopt the following metrics to measure the performance of the methods: mean Average Precision (\textbf{mAP}), Precision curves within Hamming distance 2 (\textbf{P@H}$\mathbf{=}$\textbf{2}), Precision-Recall curves (\textbf{PR}), Precision curves with different Number of top returned samples (\textbf{P@N}). For fair comparison, we adopted  MAP@5000 for CIFAR-10 and NUS-WIDE datasets and MAP@1000 for ImageNet as in~\cite{b34}.

\begin{table*}[tp]
\centering
\begin{threeparttable}
  \centering
  \caption{Configuration of the convolutional layers in ImgNets (i.e., ImgNet-$\boldsymbol{x}$, and ImgNet-$\boldsymbol{y}$).}
  \label{tab:COTCLI}
    \begin{tabular}{c|c|c|c|c|c}
    \toprule
    \specialrule{0em}{2pt}{2pt}
    \multirow{2}{*}{Layer}&
    \multicolumn{5}{c}{Configuration}\cr
    \cmidrule(lr){2-6}
    &Filter Size&Stride&Padding&LRN&Pooling\cr
    \specialrule{0em}{3pt}{3pt}
    \midrule
    \specialrule{0em}{2pt}{2pt}
    conv1&$64\ \times\ 11\ \times\ 11$&$4\ \times\ 4$&0&ON&$2\ \times\ 2$\cr
    \specialrule{0em}{1pt}{1pt}
    conv2&$256\ \times\ 5\ \times\ 5$&$1\ \times\ 1$&2&ON&$2\ \times\ 2$\cr
    \specialrule{0em}{1pt}{1pt}
    conv3&$256\ \times\ 3\ \times\ 3$&$1\ \times\ 1$&1&OFF&-\cr
    \specialrule{0em}{1pt}{1pt}
    conv4&$256\ \times\ 3\ \times\ 3$&$1\ \times\ 1$&1&OFF&-\cr
    \specialrule{0em}{1pt}{1pt}
    conv5&$256\ \times\ 3\ \times\ 3$&$1\ \times\ 1$&1&OFF&$2\ \times\ 2$\cr
    \bottomrule
    \end{tabular}
\end{threeparttable}
\end{table*}

\begin{table}[tp]
\centering
\begin{threeparttable}
  \centering
  \caption{Configuration of the fully-connected layers in ImgNets (i.e., ImgNet-$\boldsymbol{x}$, and ImgNet-$\boldsymbol{y}$).}
  \label{tab:COTFCLI}
    \begin{tabular}{c|c}
    \toprule
    \specialrule{0em}{2pt}{2pt}
    Layer&Configuration\cr
    \specialrule{0em}{3pt}{3pt}
    \midrule
    \specialrule{0em}{2pt}{2pt}
    full6&4096\cr
    \specialrule{0em}{1pt}{1pt}
    full7&4096\cr
    \specialrule{0em}{1pt}{1pt}
    Semantic layer&512\cr
    \specialrule{0em}{1pt}{1pt}
    Hash layer&$K/2$-bit hash code\cr
    \bottomrule
    \end{tabular}
\end{threeparttable}
\end{table}

\begin{table}[tp]
\centering
\begin{threeparttable}
  \centering
  \caption{Configuration of the LabelNet.}
  \label{tab:COTLN}
    \begin{tabular}{c|c}
    \toprule
    \specialrule{0em}{2pt}{2pt}
    Layer&Configuration\cr
    \specialrule{0em}{3pt}{3pt}
    \midrule
    \specialrule{0em}{2pt}{2pt}
    full-connected layer&4096\cr
    \specialrule{0em}{1pt}{1pt}
    Semantic layer&512\cr
    \specialrule{0em}{1pt}{1pt}
    Hash layer&$K/2$-bit hash code\cr
    \bottomrule
    \end{tabular}
\end{threeparttable}
\end{table}

\subsection{Implementation Details}
As shown in Figure~\ref{fig:F2}, our model consists of three networks: a LabelNet and two ImgNets. We used Alexnet~\cite{b19} for the two asymmetric ImgNets, and we add two other fully-connected layers (i.e., semantic layer and hash layer) to extract the semantic feature and project to $\mathbb{R}^{K/2}$ space, respectively. We fine-tuned convolutional layers and fully-connected layers copied from AlexNet pre-trained on ImageNet and trained the semantic layer and hashing layer by back-propagation (BP). More specifically, the overall model structure contains 5 convolutional layers (i.e., ``conv1''-``conv5'') and 4 fully-connected layers (i.e., ``full6''-``full7''-``semantic layer''-``hash layer''). The detailed configuration of the 5 convolutional layers is shown in Table~\ref{tab:COTCLI}, where ``filter size'' denotes the number of convolutional filters. ``stride'' denotes the convolutional stride. ``padding'' indicates the number of pixels to add to each size of the input feature. ``LRN'' denotes whether Local Response Normalization (LRN)~\cite{b19} is applied or not. ``pooling'' denotes the down-sampling operation. The configuration of the 4 full-connected layers is shown in Table~\ref{tab:COTFCLI}, where the numbers in the table represent the number of nodes in each layer. The LabelNet contains 3 layers, the detailed configuration of the 3 layers is shown in Table~\ref{tab:COTLN}. In our proposed \textbf{ADSQ} method, images in batch form are used as the input and every two images in the same batch constitute an image pair. The parameters of \textbf{ADSQ} model are learned by alternative training strategy. We summarize the whole learning algorithm for \textbf{ADSQ} in Algorithm 1.

\textbf{Network Parameters} In our \textbf{ADSQ}, the value of hyper-parameters are $\alpha=\beta=1$, $\gamma=10^{-2}$ and $\nu=\eta=10$. Our model is implemented on Pytorch\footnote{https://pytorch.org/} on a server with a NVIDIA TITAN X GPUs. The network is optimized by stochastic gradient descent with learning rate from $10^{-5}$ to $10^{-2}$ with a multiplicative step-size $10^{\frac{1}{2}}$. The batch size of LabelNet and two asymmetric ImgNets are set to 32 and the weight decay parameter selected was 0.0005. The momentum is set to 0.9.


\begin{table*}[tp]
\footnotesize
\centering
\begin{threeparttable}
  \centering
  \caption{mean Average Precision (mAP) of Hamming Ranking for Different Number of Bits on the Three Image Datasets.}
  \label{tab:the_overall_performance_mAP}
    \begin{tabular}{l|cccc|cccc|cccc}
    \toprule
    \specialrule{0em}{2pt}{2pt}
    \multirow{2}{*}{Method}&
    \multicolumn{4}{c|}{CIFAR-10}&\multicolumn{4}{c|}{NUS-WIDE}&\multicolumn{4}{c}{ImageNet}\cr
    \cmidrule(lr){2-5}\cmidrule(lr){6-9}\cmidrule(lr){10-13}
    &12 bits&24 bits&32 bits&48 bits&12 bits&24 bits&32 bits&48 bits&12 bits&24 bits&32 bits&48 bits\cr
    \specialrule{0em}{3pt}{3pt}
    \midrule
    \specialrule{0em}{2pt}{2pt}
    SH~\cite{b7}&0.127&0.128&0.126&0.129&0.454&0.406&0.405&0.400&0.185&0.273&0.328&0.395\cr
    \specialrule{0em}{1pt}{1pt}
    ITQ~\cite{b8}&0.162&0.169&0.172&0.175&0.452&0.468&0.472&0.477&0.305&0.363&0.462&0.517\cr
    \specialrule{0em}{1pt}{1pt}
    \hline
    \specialrule{0em}{1pt}{1pt}
    SDH~\cite{b15}&0.285&0.329&0.341&0.356&0.568&0.600&0.608&0.637&0.253&0.371&0.455&0.525\cr
    \specialrule{0em}{1pt}{1pt}
    KSH~\cite{b9}&0.303&0.337&0.346&0.356&0.556&0.572&0.581&0.588&0.136&0.233&0.298&0.342\cr
    \specialrule{0em}{1pt}{1pt}
    \hline
    \specialrule{0em}{1pt}{1pt}
    DHN~\cite{b16}&0.555&0.594&0.603&0.621&0.708&0.735&0.748&0.758&0.269&0.363&0.461&0.530\cr
    \specialrule{0em}{1pt}{1pt}
    CNNH~\cite{b17}&0.429&0.511&0.509&0.522&0.611&0.618&0.625&0.608&0.237&0.364&0.450&0.525\cr
    \specialrule{0em}{1pt}{1pt}
    DNNH~\cite{b33}&0.552&0.566&0.558&0.581&0.674&0.697&0.713&0.715&0.219&0.372&0.461&0.530\cr
    \specialrule{0em}{1pt}{1pt}
    DPSH~\cite{b51}&0.713&0.727&0.744&0.757&0.752&0.790&0.794&0.812&0.143&0.268&0.304&0.407\cr
    \specialrule{0em}{1pt}{1pt}
    DSDH~\cite{b34}&0.726&0.762&0.785&0.803&0.743&0.782&0.799&0.816&0.312&0.353&0.481&0.533\cr
    \specialrule{0em}{1pt}{1pt}
    \hline
    \specialrule{0em}{1pt}{1pt}
    DSEH~\cite{bdseh}&0.753&0.781&0.807&0.822&0.745&0.785&0.811&0.819&0.449&0.487&0.545&0.576\cr
    \specialrule{0em}{1pt}{1pt}
    \textbf{ADSQ}&\textbf{0.792}&\textbf{0.823}&\textbf{0.836}&\textbf{0.851}&\textbf{0.761}&\textbf{0.793}&\textbf{0.828}&\textbf{0.833}&\textbf{0.493}&\textbf{0.553}&\textbf{0.621}&\textbf{0.649}\cr
    \specialrule{0em}{2pt}{2pt}
    \bottomrule
    \end{tabular}
\end{threeparttable}
\end{table*}

\begin{figure*}
 \centering
  \subfigure[Precision-Recall curve @ 48 bits]{
   \includegraphics[width=2.15in]{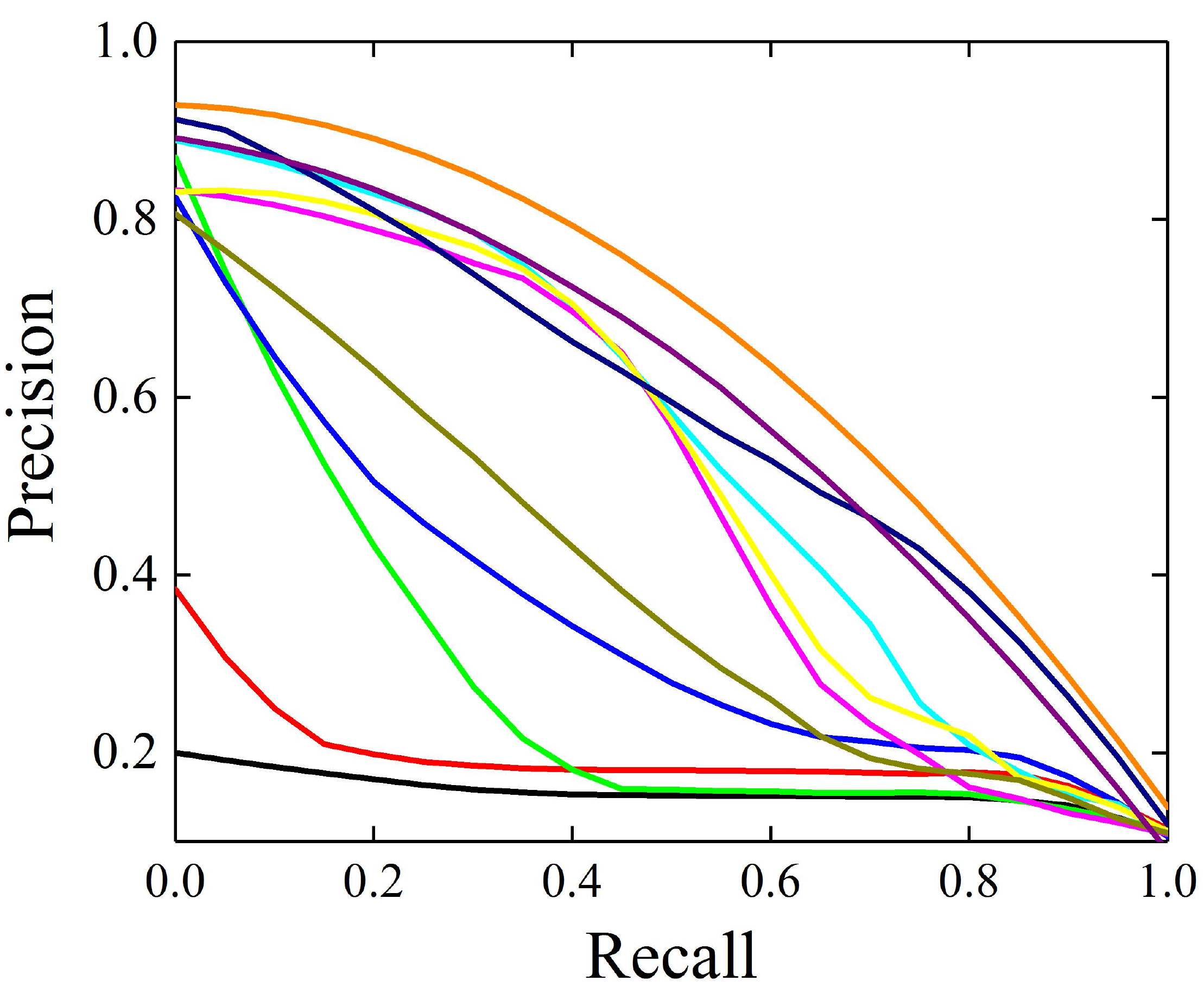}
   }
  \subfigure[Precision curve w.r.t. top-$n$ @ 48 bits]{
   \includegraphics[width=2.13in]{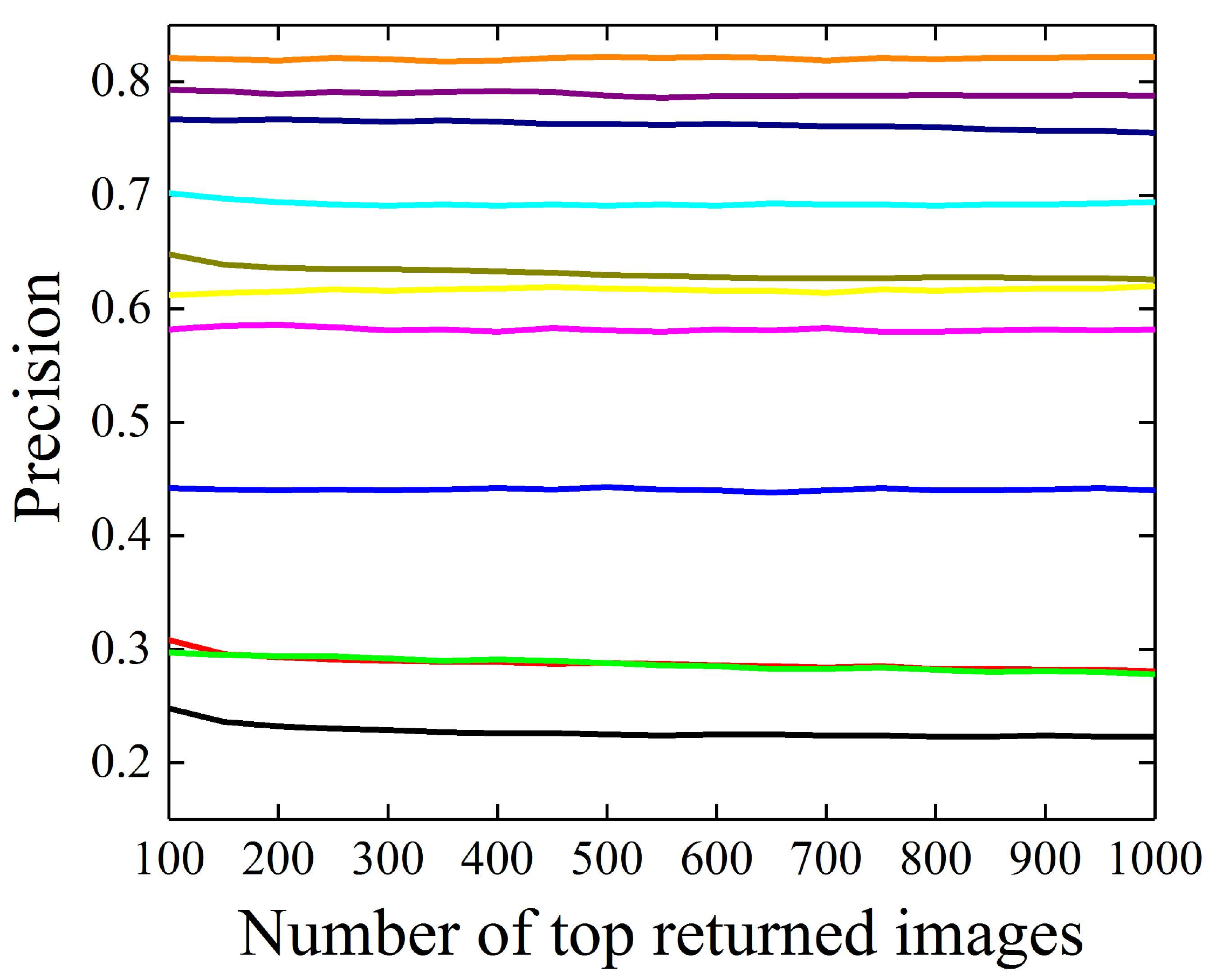}
 }
  \subfigure[Precision within Hamming radius 2]{
 \includegraphics[width=2.431in]{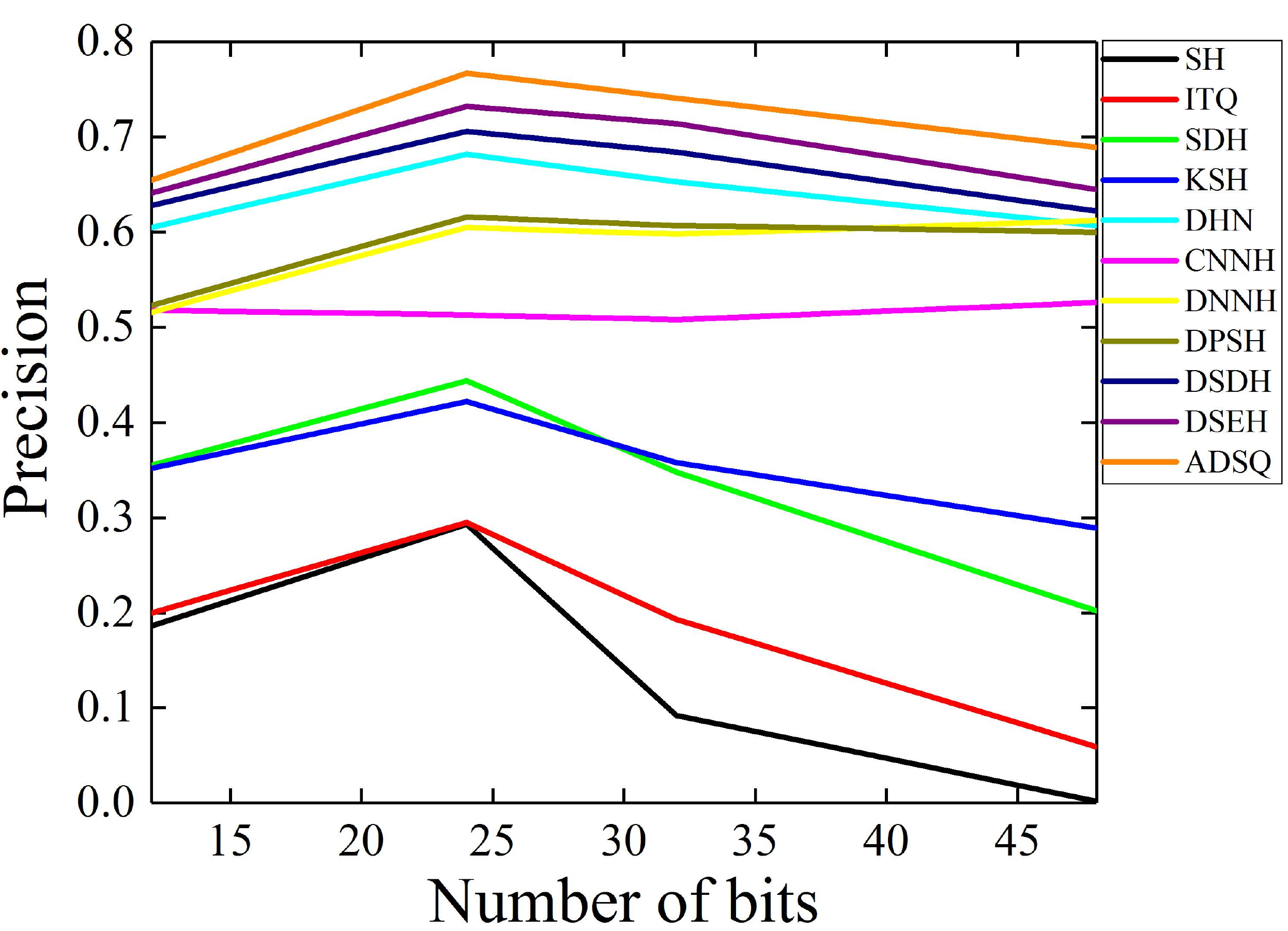}
 }

 \caption{The results of \textbf{ADSQ} and comparison methods on the CIFAR-10 dataset under three evaluation metrics.}
 \label{fig:C10} 
 \end{figure*}

\begin{figure*}
 \centering
  \subfigure[Precision-Recall curve @ 48 bits]{
   \includegraphics[width=2.148in]{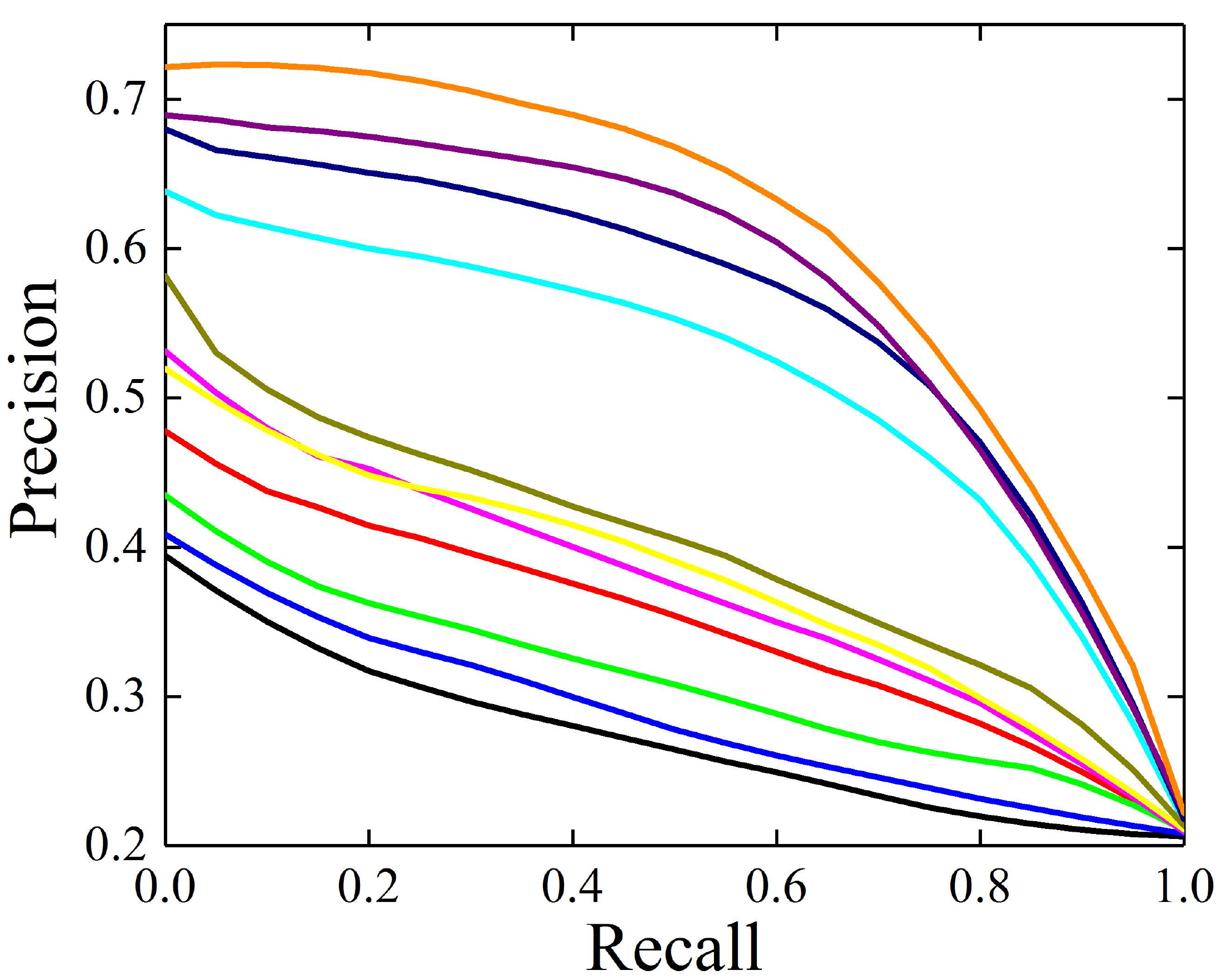}
   }
   \subfigure[Precision curve w.r.t. top-$n$ @ 48 bits]{
   \includegraphics[width=2.136in]{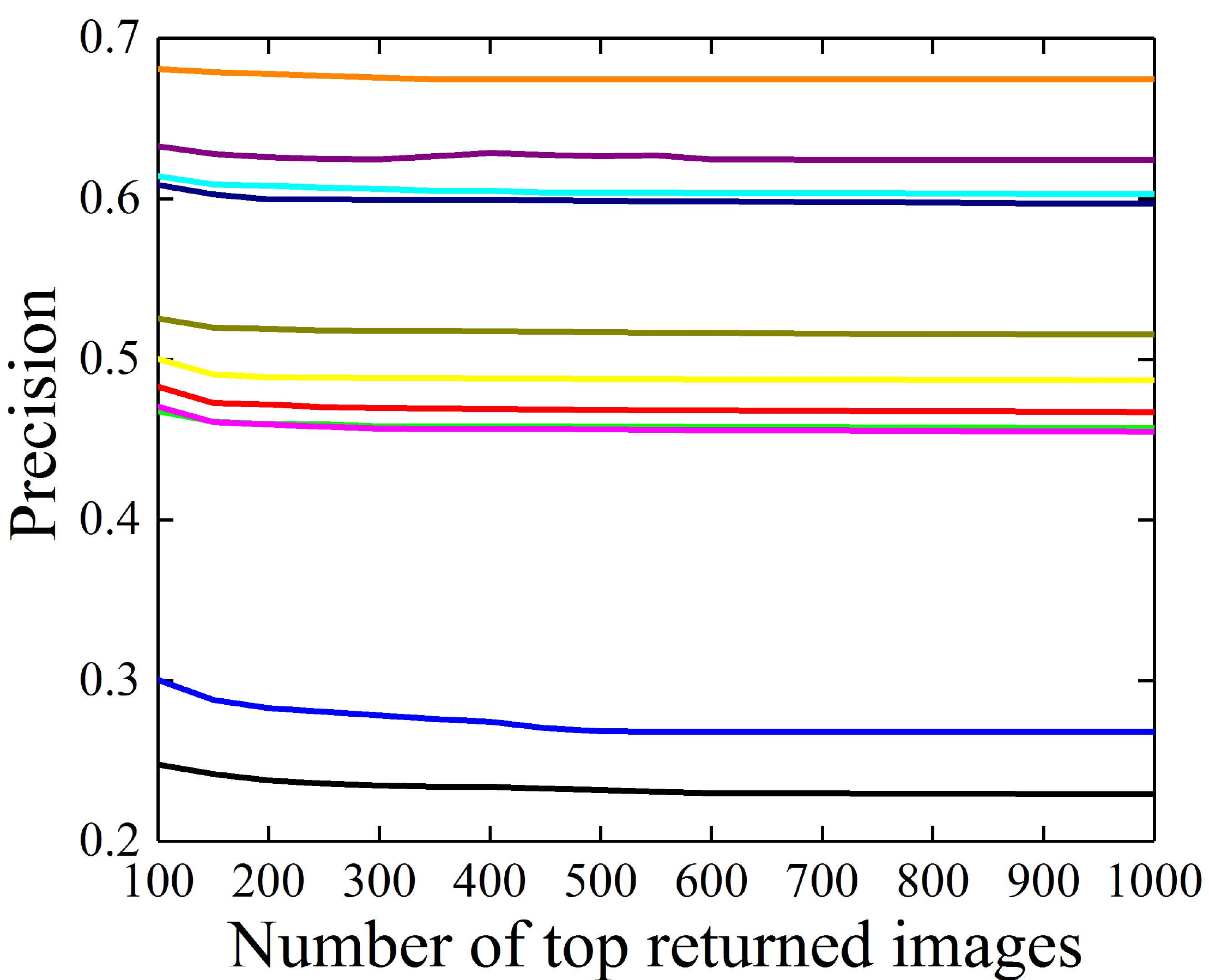}
   }
  \subfigure[Precision within Hamming radius 2]{
 \includegraphics[width=2.433in]{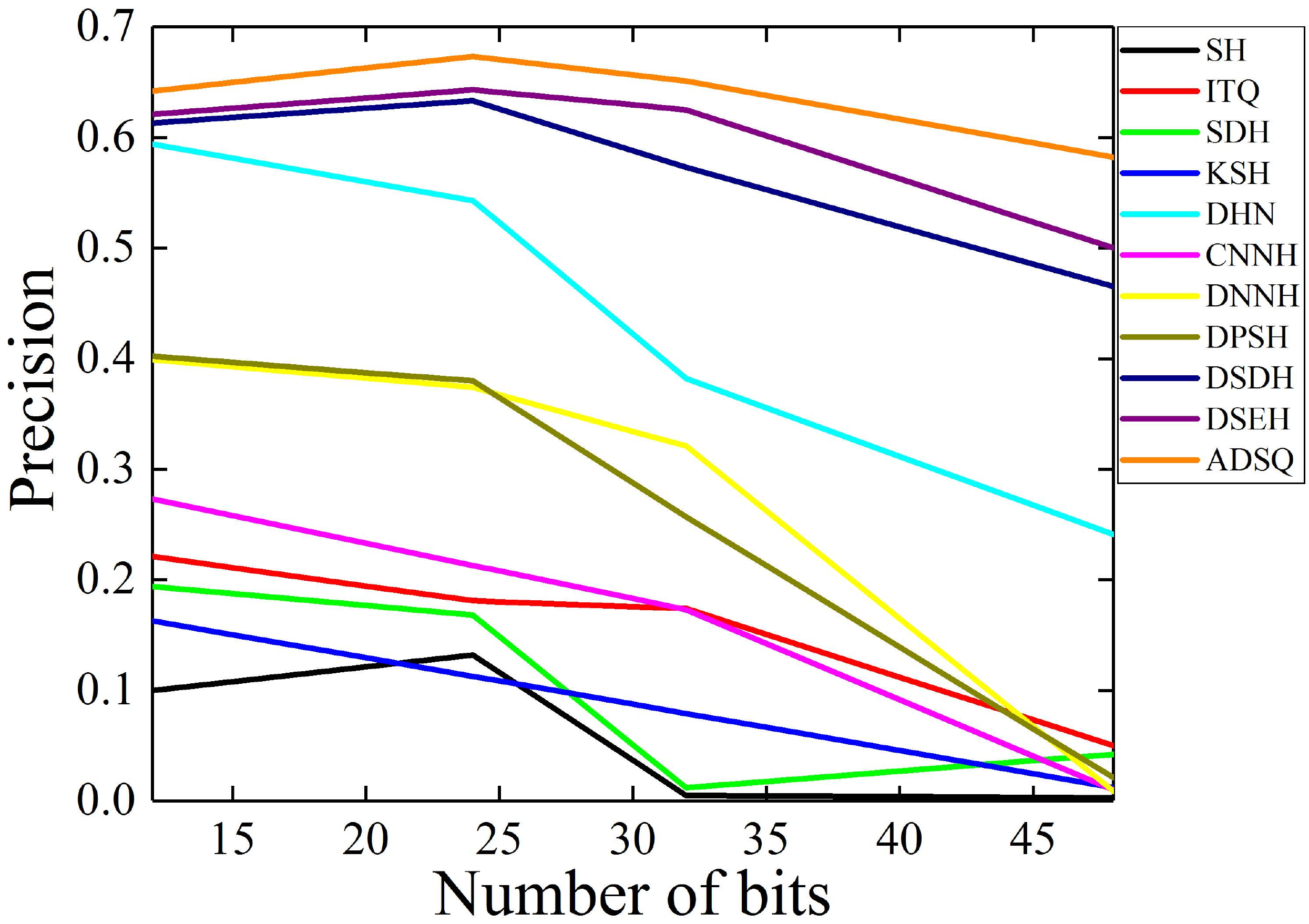}
 }

 \caption{The results of \textbf{ADSQ} and comparison methods on the NUS-WIDE dataset under three evaluation metrics.}
 \label{fig:NUS_WIDE} 
 \end{figure*}

\begin{figure*}
 \centering
  \subfigure[Precision-Recall curve @ 48 bits]{
   \includegraphics[width=2.1306in]{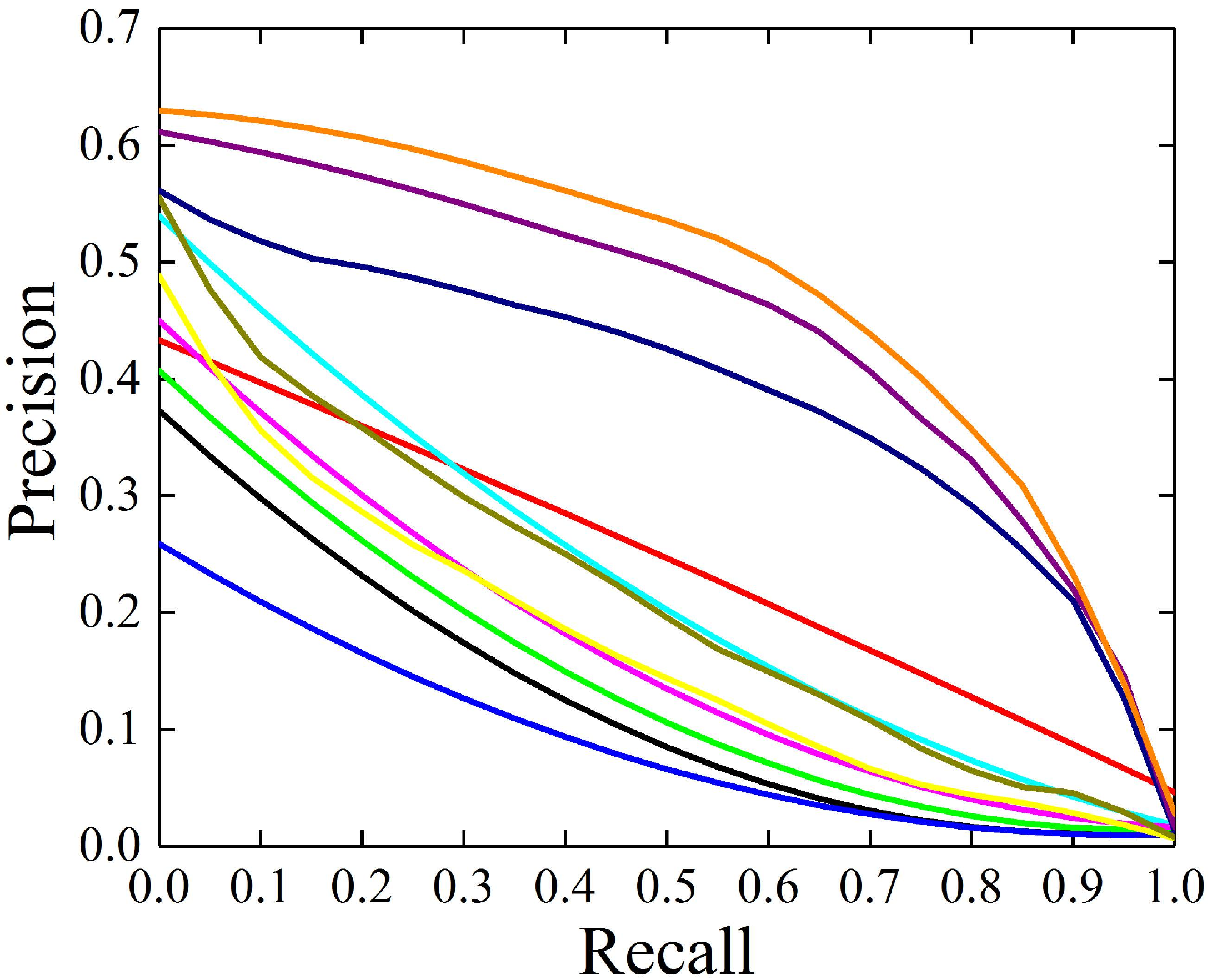}
   }
   \subfigure[Precision curve w.r.t. top-$n$ @ 48 bits]{
   \includegraphics[width=2.1454in]{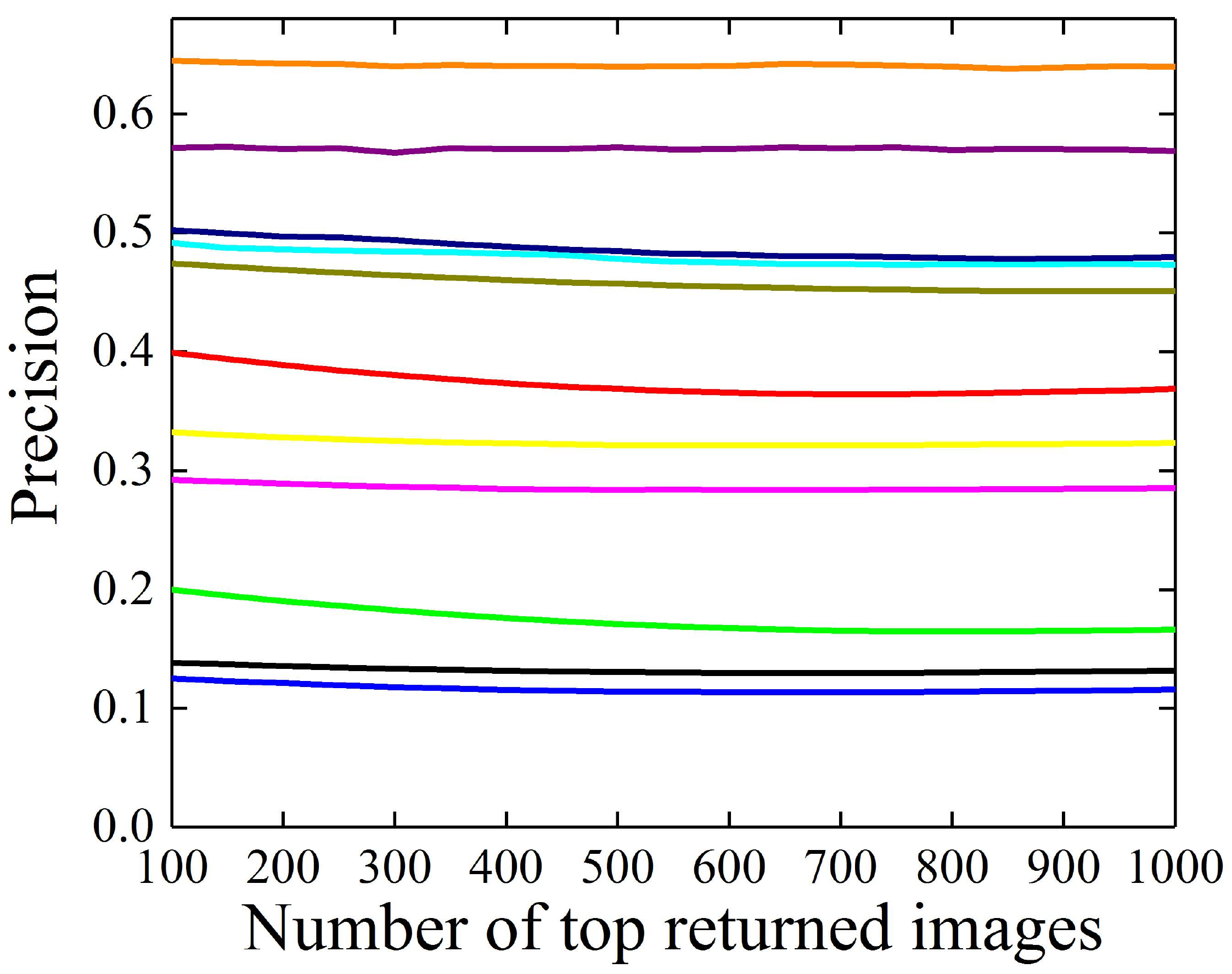}
   }
  \subfigure[Precision within Hamming radius 2]{
 \includegraphics[width=2.442in]{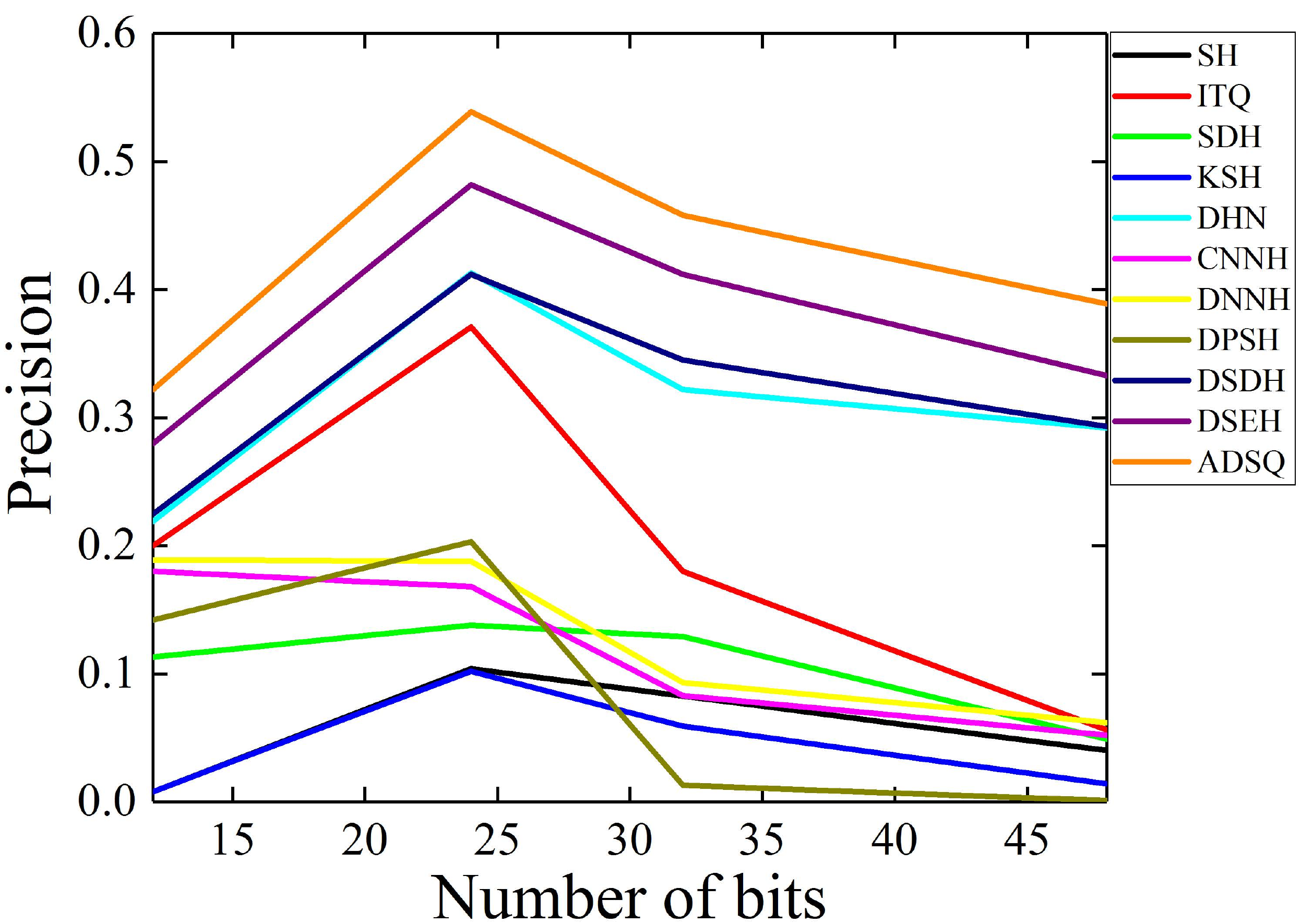}
 }

 \caption{The results of \textbf{ADSQ} and comparison methods on the ImageNet dataset under three evaluation metrics.}
 \label{fig:Image} 
 \end{figure*}

\subsection{Results and Discussions}
Table~\ref{tab:the_overall_performance_mAP} reports the mAP results on CIFAR-10, NUS-WIDE, and ImageNet dataset, respectively. The length of the hash codes varies from 12 to 48 (i.e., 12, 24, 36, and 48). From the Table~\ref{tab:the_overall_performance_mAP}, it can be observed that the performance of our \textbf{ADSQ} achieves the best image retrieval accuracy, and \textbf{ADSQ} is better than all baseline methods, including unsupervised hashing methods, supervised hashing methods, learning based hashing methods, and semantic supervised learning based hashing methods. Specifically, compared to the best shallow hashing method (i.e., ITQ) using deep features achieves an absolute score of more than 78\% increase on the mAP performance measure metric for image retrieval on the CIFAR-10 dataset. Compared to the best learning based hashing method, i.e., DSDH, our \textbf{ADSQ} achieves an absolute score of more than 8\% increase on the mAP performance measure metric. When comparing our \textbf{ADSQ} with the semantic supervised learning based hashing method DSEH, it can be seen that \textbf{ADSQ} can achieve a more than 3\% increase in mAP. On the multi-label dataset NUS-WIDE, compared to the best shallow hashing method, i.e., ITQ, our \textbf{ADSQ} achieves an absolute score of more than a 40\% increase in mAP. Compared to the best learning based hashing method, i.e., DSDH, our \textbf{ADSQ} achieves an absolute score of more than a 3\% increase in mAP. When compared to the semantic supervised hashing method, i.e., DSEH, our \textbf{ADSQ} achieves an absolute score of more than a 1.5\% increase in mAP. On large-scale dataset ImageNet, compared with ITQ, DSDH, and DSEH, our \textbf{ADSQ} achieves an absolute score of more than a 20\%, 17\%, and 8\% increase in mAP, respectively. The main difference between our proposed \textbf{ADSQ} and DSEH is that our \textbf{ADSQ} utilizes semantic information to guide the asymmetric discrete learning procedure but DSEH does not have an asymmetric structure to generate the discriminative compact hash codes. Therefore, the results demonstrate that the motivation of \textbf{ADSQ}, i.e., using semantic information to guide the asymmetric discrete learning procedure can improve image retrieval performance in practical applications. Through an in-depth analysis of Table~\ref{tab:the_overall_performance_mAP}, we can find some other insights. \textbf{(1)} By comparing KSH, SDH to SH, we can observe that the supervised hashing methods can outperform unsupervised hashing methods because the supervised information can improve performance. \textbf{(2)} By comparing DSDH, DPSH, DNNH, CNNH, DHN to SDH, we find that the learning based hashing methods can significantly outperform the traditional hashing methods. These results demonstrate the advantages of using a deep end-to-end learning structure. \textbf{(3)} By comparing semantic supervised learning based hashing methods, i.e., \textbf{ADSQ} and DSEH to other baseline hashing methods, we can find that semantic learning based deep hashing can outperform similar learning based deep hashing methods, which means that by using semantic information we are able to learn more optimal binary codes. \textbf{(4)} The performance of all methods keeps improving with the increase in hash code length.

As shown in Figures~\ref{fig:C10}(a),~\ref{fig:NUS_WIDE}(a),~\ref{fig:Image}(a) and Figures~\ref{fig:C10}(b),~\ref{fig:NUS_WIDE}(b),~\ref{fig:Image}(b), experiments were conducted to evaluate the performance by using the metrics of Precision-Recall curves (\textbf{PR}) and Precision curves with a different Number of top returned samples (\textbf{P@N}), respectively. These metrics are widely used in deploying practical applications. The proposed \textbf{ADSQ} method significantly outperforms all the baseline methods it was compared to. In particular, \textbf{ADSQ} achieves higher precision at lower recall levels and smaller number of top returned images than all compared baseline methods. This is very important for image retrieval precision as the primary purpose, where it takes only a small $N$ to count more on the top-$N$ returned results. This proves the value of the \textbf{ADSQ} method in actual image retrieval systems.

The other important indicator is Precision within Hamming radius 2 (\textbf{P@H=2}) since it only requires $O(1)$ time for each query operation. As shown in Figures~\ref{fig:C10}(c),~\ref{fig:NUS_WIDE}(c), and~\ref{fig:Image}(c), \textbf{ADSQ} achieves the highest \textbf{P@H=2} results on all the datasets with regards to different hash code lengths. This validates the assertion that the proposed \textbf{ADSQ} method can attain higher-quality hash codes than all baseline methods and can enable more efficient and accurate Hamming space retrieval. When the length of the hash codes increases, few data points fall within the Hamming sphere with a radius of 2, which is caused by the sparse of the Hamming space~\cite{b62}. Therefore, many learning based hashing methods can achieve good image retrieval performances on short hash codes. It is worth noting that \textbf{ADSQ} achieves a relatively slightly decrease in accuracy by longer code lengths, validating that \textbf{ADSQ} can effectively concentrate hash codes of similar data points together to be within the Hamming radius 2.

\begin{table}[tp]
\centering
\begin{threeparttable}
  \centering
  \caption{The mAP results of ablation study of our \textbf{ADSQ} on NUS-WIDE dataset.}
  \label{tab:abstudy}
    \begin{tabular}{ccccc}
    \toprule
    Method&12 bits&24 bits&32 bits&48 bits\cr
    \midrule
    \specialrule{0em}{2pt}{2pt}
    \textbf{ADSQ}&\textbf{0.761}&\textbf{0.793}&\textbf{0.828}&\textbf{0.833}\cr
    \specialrule{0em}{2pt}{2pt}
    \textbf{ADSQ}-$sym$&0.756&0.790&0.817&0.824\cr
    \specialrule{0em}{2pt}{2pt}
    \textbf{ADSQ}-$\mathcal{A}$&0.651&0.677&0.729&0.740\cr
    \specialrule{0em}{2pt}{2pt}
    \textbf{ADSQ}-$\mathcal{S}$&0.743&0.774&0.811&0.817\cr
    \specialrule{0em}{2pt}{2pt}
    \textbf{ADSQ}-$\mathcal{AS}$&0.637&0.651&0.699&0.714\cr
    \bottomrule
    \end{tabular}
\end{threeparttable}
\end{table}

\begin{figure*}
 \centering
  \subfigure[$\alpha$]{
   \includegraphics[width=2in]{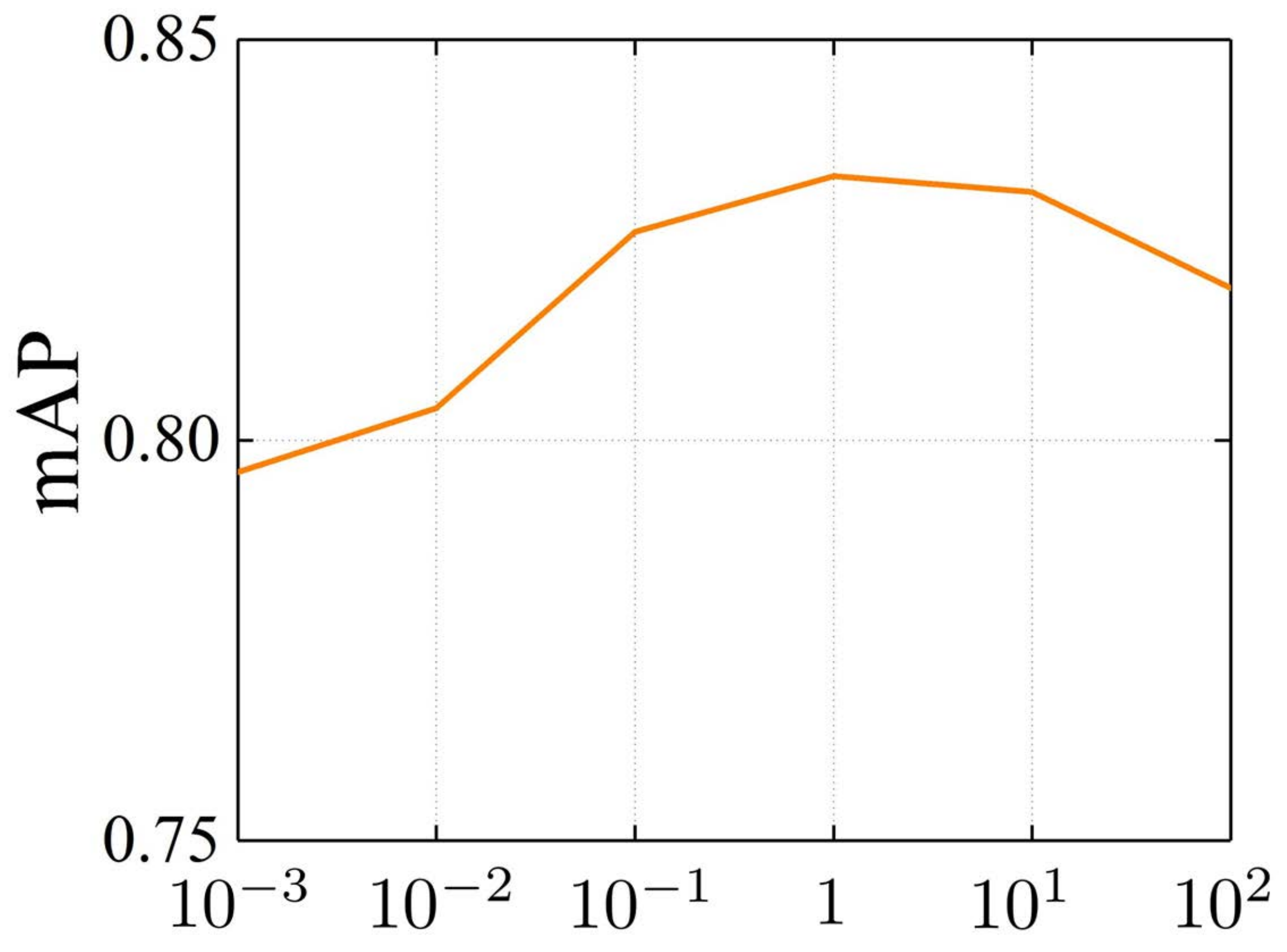}
   }
  \subfigure[$\beta$]{
 \includegraphics[width=2in]{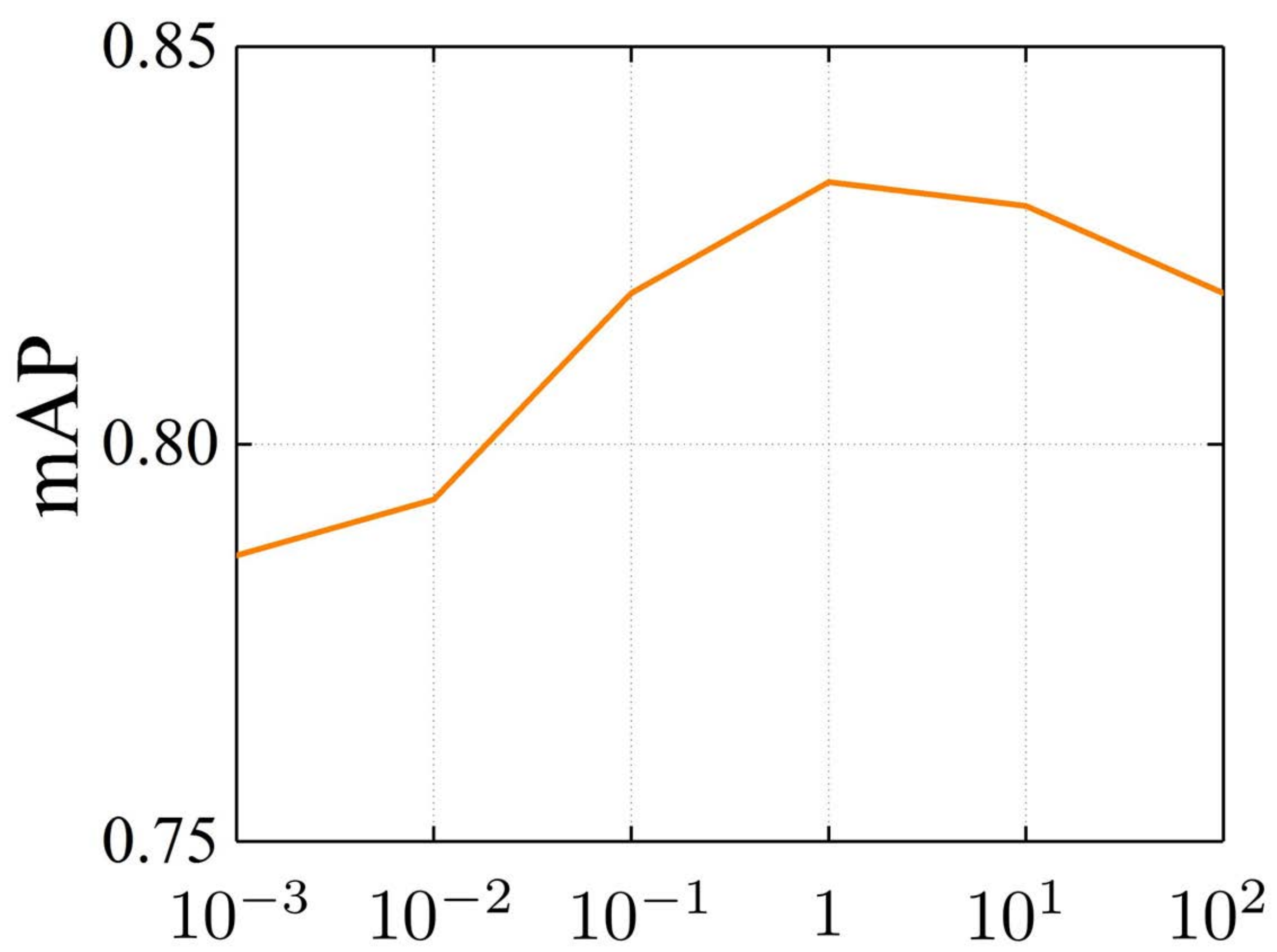}
 }
  \subfigure[$\gamma$]{
   \includegraphics[width=2in]{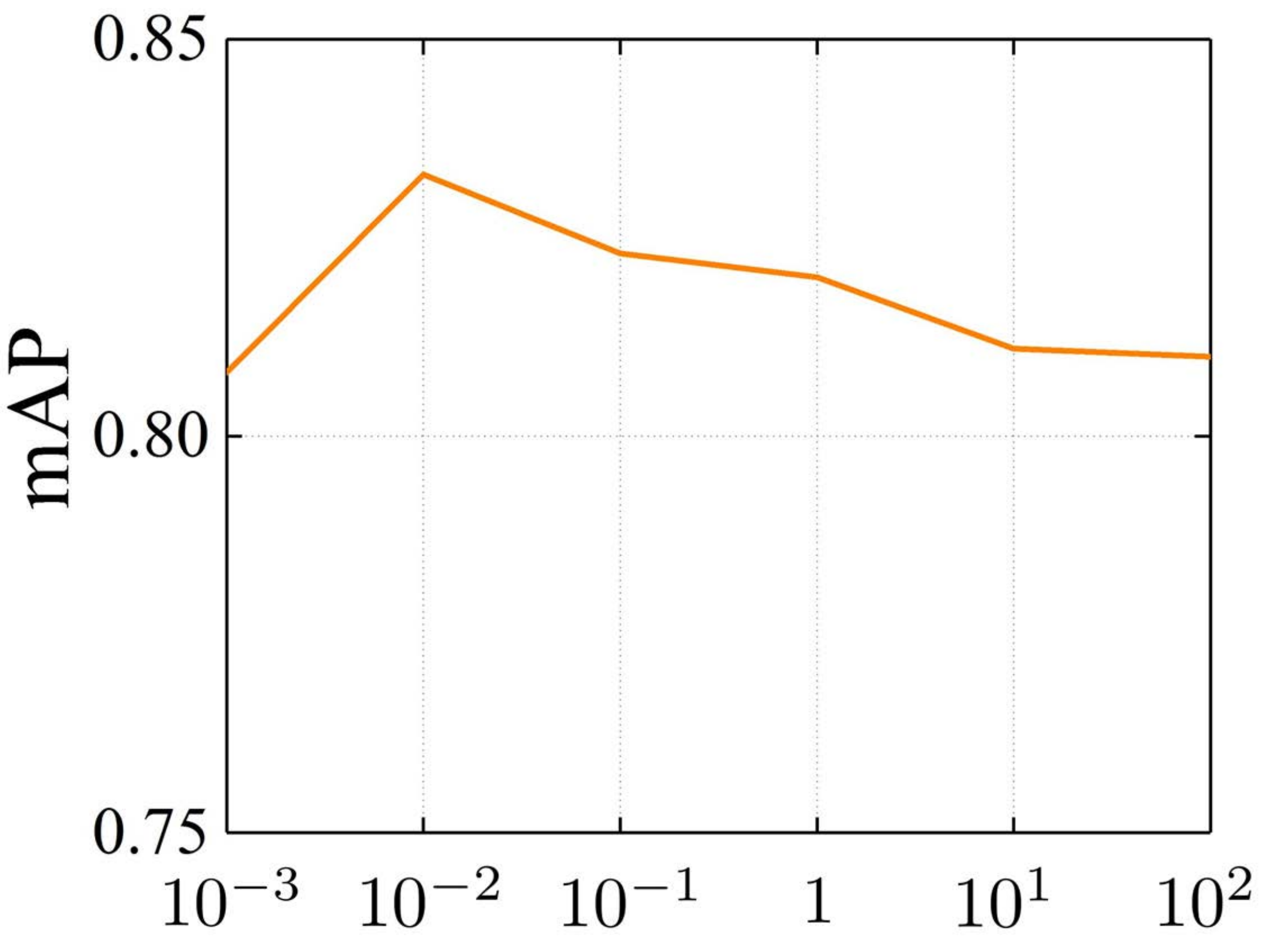}
 }
 \subfigure[$\nu$]{
   \includegraphics[width=2in]{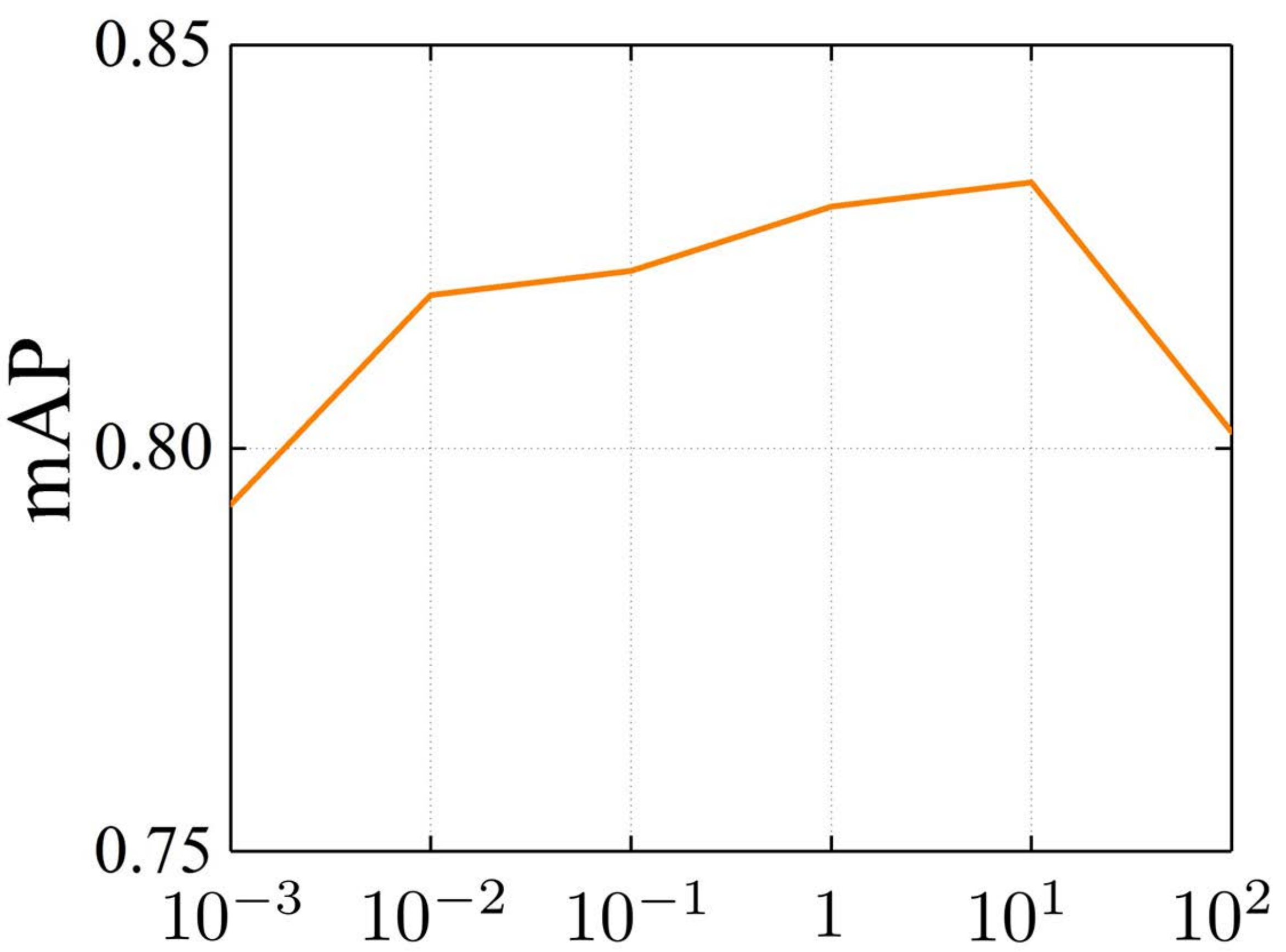}
 }
 \subfigure[$\eta$]{
   \includegraphics[width=2in]{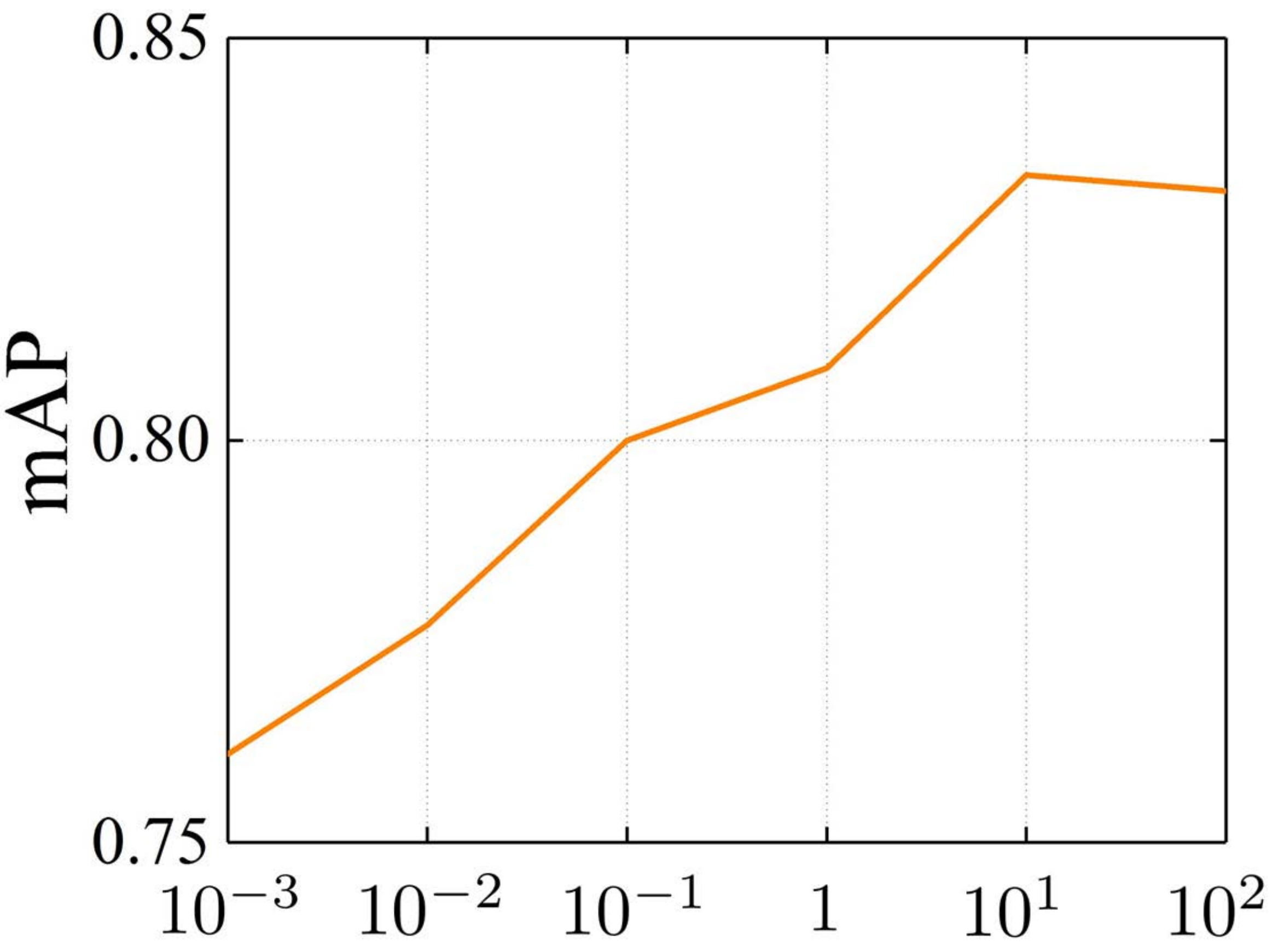}
 }
 \caption{A sensitivity analysis of the hyper-parameters (i.e., $\alpha$, $\beta$, $\gamma$, $\nu$, $\eta$).}
 \label{fig:hyperparameters} 
 \end{figure*}

\begin{figure}
 \centering
  \subfigure[ADSQ]{
  \label{fig:v:a} 
   \includegraphics[width=1.59in]{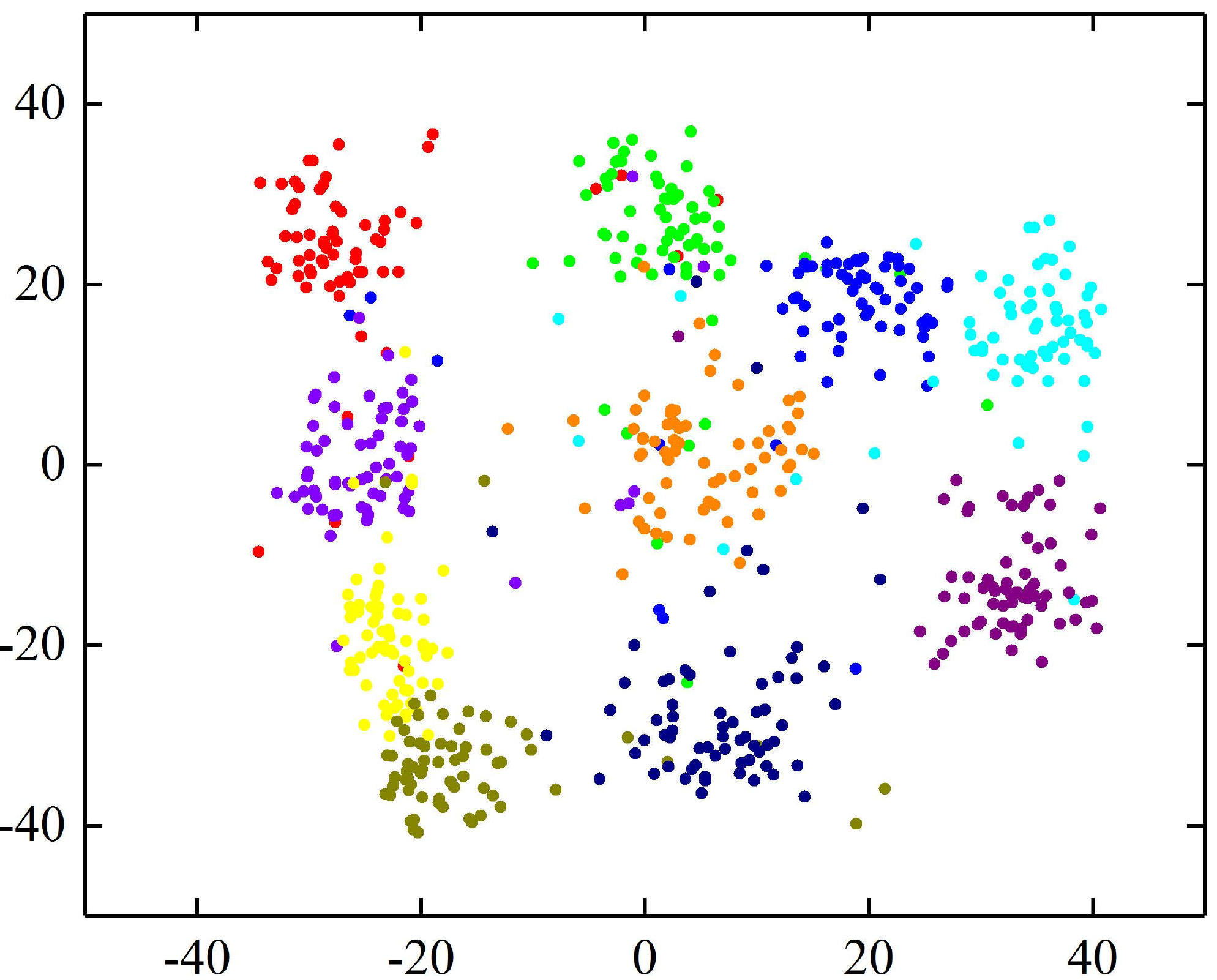}
   }
  \subfigure[DSEH]{
  \label{fig:v:b} 
 \includegraphics[width=1.59in]{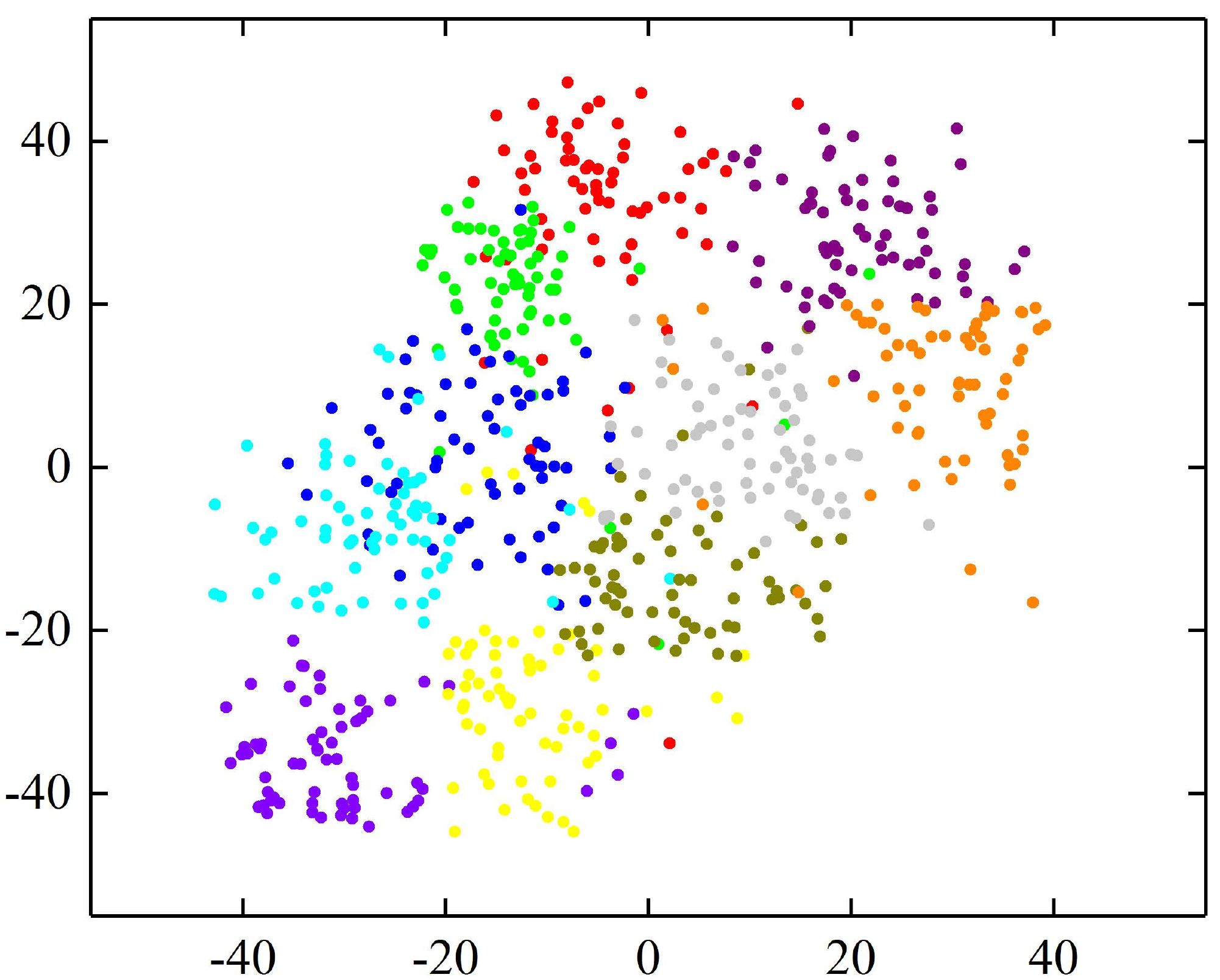}
 }
 \caption{The t-SNE visualization of hash codes learned by \textbf{ADSQ} and DSEH.}
 \label{fig:v} 
 \end{figure}

\subsection{Discussion}
\subsubsection{Ablation Study}
\textbf{In this section we will analyze the role of the asymmetric loss term $\mathcal{A}$ and semantic supervision $\mathcal{J}_1$ of~\eqref{eq:ImgLoss}, and the reason for choosing an asymmetric structure}.

In order to demonstrate that the asymmetric loss term and semantic supervision of ~\eqref{eq:ImgLoss} and the asymmetric structure are necessary for \textbf{ADSQ}, we designed four variants of \textbf{ADSQ} on the NUS-WIDE dataset. \textbf{ADSQ}-$\mathcal{A}$ denotes a variant where~\eqref{eq:ImgLoss} is used without the asymmetric loss term. Therefore, the $\mathcal{L}^\kappa_\mathcal{A}$ can be rewritten as $\mathcal{L}^\kappa_\mathcal{A}= \alpha\mathcal{J}_1+\beta\mathcal{J}_2+\eta\mathcal{J}_3+\nu\mathcal{J}_4$. \textbf{ADSQ}-$\mathcal{S}$ denotes a variant of \textbf{ADSQ} where~\eqref{eq:ImgLoss} is used without the semantic supervision loss term. Thus, the $\mathcal{L}^\kappa_\mathcal{S}$ can be rewritten as to $\mathcal{L}^\kappa_\mathcal{S}= \beta\mathcal{J}_2+\eta\mathcal{J}_3+\nu\mathcal{J}_4+\mathcal{A}$. In the variant \textbf{ADSQ}-$\mathcal{AS}$, ~\eqref{eq:ImgLoss} is used without both the asymmetric loss term and the semantic supervision. Thus, the $\mathcal{L}^\kappa_\mathcal{AS}$ can be rewritten as $\mathcal{L}^\kappa_\mathcal{AS}=\beta\mathcal{J}_2+\eta\mathcal{J}_3+\nu\mathcal{J}_4$. The final variant of \textbf{ADSQ} is called \textbf{ADSQ}-$sym$ where we use the symmetric structure (uses the same convolutional neural network to generate the compact hash codes like DSEH~\cite{bdseh}). The mAP results are shown in Table~\ref{tab:abstudy}. From Table~\ref{tab:abstudy} the following observations were made:
\begin{enumerate}
  \item \textbf{ADSQ} outperforms \textbf{ADSQ}-$\mathcal{A}$, \textbf{ADSQ}-$\mathcal{S}$, and \textbf{ADSQ}-$\mathcal{AS}$ on all cases on NUS-WIDE dataset, which confirms the assertion that the asymmetric loss term $\mathcal{A}$ and semantic supervision term $\mathcal{J}_1$ are necessary for \textbf{ADSQ}.
  \item The gap between \textbf{ADSQ} and \textbf{ADSQ}-$\mathcal{A}$ is larger than that between \textbf{ADSQ} and \textbf{ADSQ}-$\mathcal{S}$. This result demonstrates that the asymmetric loss term $\mathcal{A}$ has a greater impact on \textbf{ADSQ} than the semantic supervision term $\mathcal{J}_1$.
  \item The performance with asymmetric structure (\textbf{ADSQ}) is better than the symmetric one (\textbf{ADSQ}-$sym$). The reason is that the use of a symmetric structure usually leads to highly correlated bits in practice, limiting the performance of image retrieval, and the use of the asymmetric structure to learn half of the codes can reduce the correlation between hash codes and enhance the robustness of the learned hash codes.
\end{enumerate}

\subsubsection{Sensitivity Analysis}
In this subsection, we analyze the impact of the hyper-parameters, i.e., $\alpha,\ \beta,\ \gamma,\ \nu$, and $\eta$. The experiments are conducted on the NUS-WIDE dataset. We tune a hyper-parameter with others fixed. Specifically, we tune $\alpha$ by fixing $\beta=1$, $\gamma=10^{-2}$ and $\nu=\eta=10$. Similarly, we fix  $\alpha=1$, $\gamma=10^{-2}$ and $\nu=\eta=10$ when tuning the value of $\beta$ and so on. As shown in Figure~\ref{fig:hyperparameters}, our model is not affected much by the change of hyper-parameters. This results demonstrate the robustness of our proposed method.

\subsubsection{Visualization}
To better illustrate the discriminative ability of \textbf{ADSQ}, the distribution of the hash codes learned by the proposed \textbf{ADSQ} method and the state-of-the-art semantic supervised hashing method DSEH on the ImageNet dataset with 48 bits are visualized by using t-SNE visualization~\cite{b63} (we sample 10 categories for the case of visualization). It can be observed that the learned hash codes by the proposed \textbf{ADSQ} method are more discriminative than those learned by DSEH. That is, the learned hash codes by \textbf{ADSQ} are more discriminative.

\section{Conclusion}\label{sec:Conclusion_and_Future_Work}
In this work, we proposed a novel supervised hashing approach, dubbed \textbf{\underline{A}}symmetric \textbf{\underline{D}}eep \textbf{\underline{S}}emantic \textbf{\underline{Q}}uantization (\textbf{ADSQ}), for image retrieval. \textbf{ADSQ} consists a LabelNet and two asymmetric \emph{ImgNets}, the LabelNet is used to discover semantic information from labels. The two asymmetric \emph{ImgNets} are used to generate their respective discriminative compact hash codes. Moreover, \textbf{ADSQ} uses rich semantic information to guide the two \emph{ImgNets} in minimizing the gap between the real-continuous features and discrete binary codes. \textbf{ADSQ} is the first asymmetric supervised hashing method which can use the abundant semantic information generated by \emph{LabelNet} to guide the discrete hash code generation of asymmetric \emph{ImgNets}. Extensive experiments on the three benchmark datasets demonstrated that the proposed \textbf{ADSQ} achieves the best performance in contrast with several state-of-the-art methods. In the future, we will use two asymmetric networks with different structures to generate high-quality hash codes.


\section*{Acknowledgment}

The authors would also like to thank the associate editor and anonymous reviewers for their comments to improve the paper.



\begin{thebibliography}{00}

\bibitem{bNNS} A. Andoni, ``Nearest Neighbor Search in High-dimensional Spaces,'' in \emph{Proceedings of International Symposium Mathematical Foundations of Computer Science (MFCS)}, Aug. 2011, pp. 1-33.

\bibitem{b1} Q. Y. Jiang, X. Cui, and W. J. Li, ``Deep Discrete Supervised Hashing,'' \emph{IEEE Trans. Image Processing}, vol. 27, no. 12, pp. 5996-6009, 2018.

\bibitem{b2} Z. K. Chen, F. M. Zhong, G. Y. Min, Y. L. Leng, and Y. M. Ying, ``Supervised Intra- and Inter-Modality Similarity Preserving Hashing for Cross-Modal Retrieval,'' IEEE Access, vol.6, pp. 27796-27808, 2018.

\bibitem{b3} Q. Y. Jiang, and W. J. Li, ``Asymmetric Deep Supervised Hashing,'' in \emph{Proceedings of the Conference on Artificial Intelligence (AAAI)}, Feb. 2018, pp. 3342-3349.

\bibitem{bdagh} Z. Yang, O. I. Raymond, W. Q. Sun, and J. Long, ``Deep Attention-Guided Hashing,'' IEEE Access, vol. 7, pp. 11209-11221, 2019.

\bibitem{b4} T. T. Yuan, W. H. Deng, and J. N. Hu, ``Distortion Minimization Hashing,'' IEEE Access, vol. 5, pp. 23425-23435, 2017.


\bibitem{b5} C. Deng, Z. J. Chen, X. L. Liu, X. B. Gao, and D. C. Tao, ``Triplet-Based Deep Hashing Network for Cross-Modal Retrieval,'' \emph{IEEE Trans. Image Processing}, vol. 27, no. 8, pp. 3893-3903, 2018.

\bibitem{b6} H. Liu, M. B. Lin, S. C. Zhang, Y. J. Wu, F. Y. Huang, and R. R. Ji, ``Dense Auto-Encoder Hashing for Robust Cross-Modality Retrieval,'' in \emph{Proceedings of Conference on ACM Multimedia (ACMMM)}, Oct. 2018, pp. 1589-1597.


\bibitem{bisoh} W. H. Kong, and W. J. Li, ``Isotropic Hashing,'' in \emph{International Conference on Neural Information Processing (NIPS)}, Dec. 2012, pp. 1655-1663.

\bibitem{bdgh} X. S. Shi, F. Y. Xing, K. D. Xu, M. Sapkota, and L. Yang, ``Asymmetric Discrete Graph Hashing,'' in \emph{Proceedings of the Conference on Artificial Intelligence (AAAI)}, Feb. 2017, pp. 2541-2547.

\bibitem{b8} Y. C. Gong, S. Lazebnik, A. Gordo, and F. Perronnin, ``Iterative Quantization: A Procrustean Approach to Learning Binary Codes for Large-Scale Image Retrieval,'' \emph{IEEE Trans. Pattern Anal. Mach. Intell.}, vol. 35, no. 12, pp.2916-2929, 2013.

\bibitem{b9} W. Liu, J. Wang, R. R. Ji, Y. G. Jiang, and S. F. Chang, ``Supervised hashing with kernels,'' in \emph{IEEE Conference on Computer Vision and Pattern Recognition (CVPR)}, Jun. 2012, pp.2074-2081.

\bibitem{b11} M. Norouzi, D. J. Fleet, ``Minimal Loss Hashing for Compact Binary Codes,'' in \emph{Proceedings of International Conference on Machine Learning (ICML)}, Jun. 2011, pp.353-360.

\bibitem{b12} J. F. Wang, J. D. Wang, N. H. Yu, and S. P. Li, ``Order preserving hashing for approximate nearest neighbor search,'' in \emph{Proceedings of Conference on ACM Multimedia (ACMMM)}, Oct. 2013, pp.133-142.

\bibitem{bcos} W. C. Kang, W. J. Li, and Z. H. Zhou, ``Column Sampling Based Discrete Supervised Hashing,'' in \emph{Proceedings of the Thirtieth Conference on American Association for Artificial Intelligence (AAAI)}, Feb. 2016, pp. 1230-1236.

\bibitem{b15} F. Shen, C. H. Shen, W. Liu, and H. T. Shen, ``Supervised Discrete Hashing,'' in \emph{IEEE Conference on Computer Vision and Pattern Recognition (CVPR)}, Jun. 2015, pp.37-45.

\bibitem{b20} K. M. He, X. Y. Zhang, S. Q. Ren, and J. Sun, ``Deep Residual Learning for Image Recognition,'' in \emph{IEEE Conference on Computer Vision and Pattern Recognition (CVPR)}, Jun. 2016. pp. 770-778.

\bibitem{b10} J. Youn, J. Shim, and S. G. Lee, ``Efficient Data Stream Clustering With Sliding Windows Based on Locality-Sensitive Hashing,'' IEEE Access, vol. 6, pp. 63757-63776, 2018.

\bibitem{b16} H. Zhu, M. S. Long, J. M. Wang, and Y. Cao, ``Deep Hashing Network for Efficient Similarity Retrieval,'' in \emph{Proceedings of the Thirtieth Conference on American Association for Artificial Intelligence (AAAI)}, Feb. 2016, pp. 2415-2421.

\bibitem{b17} P. K. Xia, Y. Pan, H. J. Lai, C. Liu, and S. C. Yan, ``Supervised Hashing for Image Retrieval via Image Representation Learning,'' in \emph{Proceedings of the Conference on Artificial Intelligence (AAAI)}, Jul. 2014, pp. 2156-2162.

\bibitem{bwdtns} X. B. Shu, G. J. Qi, J. H. Tang, and J. D. Wang, ``Weakly-Shared Deep Transfer Networks for Heterogeneous-Domain Knowledge Propagation,'' in \emph{Proceedings of Conference on ACM Multimedia (ACMMM)}, Oct. 2015, pp. 35-44.

\bibitem{b18} J. Wu, Y. He, X. N. Guo, Y. J. Zhang, and N. Zhao, ``Heterogeneous Manifold Ranking for Image Retrieval,'' IEEE Access, vol. 5, pp. 16871-16884, 2017.

\bibitem{b32} H. M. Liu, R. P. Wang, S. G. Shan, and X. L. Chen, ``Deep Supervised Hashing for Fast Image Retrieval,'' in \emph{IEEE Conference on Computer Vision and Pattern Recognition (CVPR)}, Jun. 2016, pp. 2064-2072.

\bibitem{bdseh} N. Li, C. Li, C. Deng, X. L. Liu, and G. B. Gao, ``Deep Joint Semantic-Embedding Hashing,'' in \emph{Proceedings of International Joint Conference on Artificial Intelligence (IJCAI)}, Jul. 2018, pp. 2397-2403.

\bibitem{b21} J. D. Wang, T. Zhang, J. K. Song, N. Sebe, and H. T. Shen, ``A Survey on Learning to Hash,'' \emph{IEEE Trans. Pattern Anal. Mach. Intell.}, vol.40, no.4, pp.769-790, 2018.

\bibitem{bdoh} L. Jin, X. B. Shu, K. Li, Z. C. Li, G. J. Qi, and J. H. Tang, ``Deep Ordinal Hashing With Spatial Attention,'' \emph{{IEEE} Trans. Image Processing}, vol. 28, no. 5, pp. 2173-2186, 2019.

\bibitem{bdtns} J. H. Tang, X. B. Shu, Z. C. Li, G. J. Qi, and J. D. Wang, ``Generalized Deep Transfer Networks for Knowledge Propagation in Heterogeneous Domains,'' \emph{TOMCCAP}, vol. 12, no. 4s, pp. 68:1-68:22, 2016.

\bibitem{b31} F. Zhao, Y. Z. Huang, L. Wang, and T. N. Tan, ``Deep semantic ranking based hashing for multi-label image retrieval,'' in \emph{IEEE Conference on Computer Vision and Pattern Recognition (CVPR)}, Jun. 2015, pp.1556-1564.

\bibitem{b7} Y. Weiss, A. Torralba, and R. Fergus, ``Spectral Hashing,'' in \emph{International Conference on Neural Information Processing (NIPS)}, Dec. 2008, pp. 1753-1760.

\bibitem{b22} J. Leskovec, A. Rajaraman, and J. D. Ullman, ``Mining of Massive Datasets, 2nd Ed,'' \emph{Cambridge University Press}, 2014.

\bibitem{b23} B. Kulis and K. Grauman, ``Kernelized locality-sensitive hashing,'' \emph{IEEE Trans. Pattern Anal. Mach. Intell.}, vol. 34, no. 6, pp. 1092-1104, 2012.

\bibitem{b24} M. Datar, N. Immorlica, P. Indyk, and V.S. Mirrokni, ``Locality-sensitive hashing scheme based on p-stable distributions,'' in \emph{Proceedings of the twentieth  annual symposium on Computational geometry}, pp. 253-262, 2004.

\bibitem{b25} H. Jegou, M. Douze, and C. Schmid, ``Product Quantization for Nearest Neighbor Search,'' \emph{IEEE Trans. Pattern Anal. Mach. Intell.}, vol. 33, no. 1, pp. 117-128, 2011.

\bibitem{b26} Y. C. Gong and S. Lazebnik, ``Iterative quantization: A procrustean approach to learning binary codes,'' in \emph{IEEE Conference on Computer Vision and Pattern Recognition (CVPR)}, Jun. 2011, pp. 20-25.

\bibitem{b28} Y. Cao, H. Qi, W. R. Zhou, J. Kato, K. Q. Li, X. L. Liu, X. L. Liu, and J. Gui, ``Binary Hashing for Approximate Nearest Neighbor Search on Big Data: A Survey,'' IEEE Access, vol. 6, pp. 2039-2054, 2018.

\bibitem{b29} Y. Weiss, A. Torralbe, and R. Fergus, ``Spectral Hashing,'' in \emph{In Advances in Neural Information Processing Systems (NIPS)}, Dec. 2008, pp. 1753-1760.

\bibitem{b30} W. Liu, J. Wang, S. Kumar, and S.F. Chang, ``Hashing with Graphs,'' in \emph{Proceedings of the 28th International Conference on Machine Learning (ICML)}, Jun. 2011, pp. 1-8.

\bibitem{bmqe} X. L. Liu, B. Du, C. Deng, M. Liu, and B. Lang, ``Structure Sensitive Hashing With Adaptive Product Quantization,'' \emph{IEEE Trans. Cybernetics}, vol. 46, no. 10, pp. 2252-2264, 2016.

\bibitem{b33} H. J. Lai, Y. Pan, Y. Liu, and S. C. Yan, ``Simultaneous feature learning and hash coding with deep neural networks,'' in \emph{IEEE Conference on Computer Vision and Pattern Recognition (CVPR)}, Jun. 2015, pp. 3270-3278.

\bibitem{b34} Q. Li, Z. N. Sun, R. He, and T. N. Tan, ``Deep Supervised Discrete Hashing,'' in \emph{International Conference on Neural Information Processing (NIPS)}, Dec. 2017, pp. 2479-2488.

\bibitem{b19} A. Krizhevsky, I. Sutskever, and G. E. Hinton, ``ImageNet Classification with Deep Convolutional Neural Networks,'' in \emph{International Conference on Neural Information Processing (NIPS)}, Dec. 2012, pp. 1106-1114.

\bibitem{bj4} Q. Y. Jiang, and W. J. Li, ``Deep Cross-Modal Hashing,'' in \emph{IEEE Conference on Computer Vision and Pattern Recognition (CVPR)}, Jul. 2017, pp. 3270-3278.

\bibitem{b56} A. Krizhevsky and G. Hinton, ``learning multiple layers of features from tiny images,'' M.S. thesis, Dept. Comput. Sci., Univ. Toronto, Toronto, ON, Canada, 2009.

\bibitem{b57} T. S. Chua, J. H. Tang, R. C. Hong,  H. J. Li, Z. P. Luo, and Y. T. Zheng, ``NUS-WIDE: a real-world web image database from National University of Singapore,'' in \emph{Proceedings of International Conference on Image and Video Retrieval (CIVR)}, Jul. 2009.

\bibitem{b58} O. Russakovsky, J. Deng, H. Su, J. Krause, S. Satheesh, S. Ma, Z. H. Huang, A. Karpathy, A. Khosla, M. S. Bernstein, A. C. Berg, and F. F. Li, ``ImageNet Large Scale Visual Recognition Challenge,'' \emph{International Journal of Computer Vision (IJCV)}, vol. 115, no. 3, pp. 211-252, 2015.

\bibitem{b53} Z. J. Cao, Z. P. Sun, M. S. Long, J. M. Wang, and P. S. Yu, ``Deep Priority Hashing,'' in \emph{Proceedings of Conference on ACM Multimedia (ACMMM)}, Oct. 2018, pp. 1653-1661.

\bibitem{b51} W. J. Li, S. Wang, and W. C. Kang, ``Feature Learning Based Deep Supervised Hashing with Pairwise Labels,'' in \emph{Proceedings of International Joint Conference on Artificial Intelligence (IJCAI)}, Jul. 2016, pp. 1711-1717.

\bibitem{b59} K. Chatfield, K. Simonyan, A. Vedaldi, and A. Zisserman, ``Return of the Devil in the Details: Delving Deep into Convolutional Nets.'' [Online]. Available: https://arxiv.org/abs/1405.3531

\bibitem{b60} J. Donahue, Y. Jia, O. Vinyals, J. Hoffman, N. Zhang, E. Tzeng, and T. Darrell, ``DeCAF: A Deep Convolutional Activation Feature for Generic Visual Recognition, in \emph{Proceedings of International Conference on Machine Learning (ICML)}, Jun. 2014, pp.647-655.

\bibitem{b62} M. Norouzi, A. Punjani, and D. J. Fleet, ``Fast Exact Search in Hamming Space With Multi-Index Hashing,'' \emph{IEEE Trans. Pattern Anal. Mach. Intell.}, vol.36, no.6, pp.1107-1119, 2014.

\bibitem{b63} L. van der Maaten and G. Hinton, ``Visualizing high-dimensional data using t-sne,'' \emph{Journal of Machine Learning Research}, vol. 9, pp. 2579-2605, 2008.
\end{thebibliography}
\end{document}